\documentclass{article}

\PassOptionsToPackage{numbers, compress}{natbib}

 \usepackage[main, final]{neurips_2026}

\usepackage[utf8]{inputenc} 
\usepackage[T1]{fontenc}    
\usepackage{hyperref}       
\usepackage{url}            
\usepackage{booktabs}       
\usepackage{amsfonts}       
\usepackage{nicefrac}       
\usepackage{xcolor}         
\usepackage{graphicx}
\usepackage{subcaption}
\usepackage{wrapfig}

\usepackage{amsmath}
\usepackage{amssymb}
\usepackage{mathtools}
\usepackage{amsthm}
\usepackage{kotex}
\usepackage{pifont}

\theoremstyle{plain}

\theoremstyle{definition}

\theoremstyle{remark}

\usepackage{multicol, multirow}
\usepackage{array}
\usepackage{tabularx}
\usepackage{makecell}
\usepackage{subcaption}
\usepackage{stfloats}

\usepackage{algorithm}
\usepackage{algorithmic}
\newcommand{\rcomment}[1]{\hfill{\footnotesize\{#1\}}}

\title{SBNO : Schr\"odinger Bridge Neural Operator\\for the Forward and Inverse Problems \\under Missing Information}

\newcommand{\affilmark}[1]{\textsuperscript{#1}}

\author{%
\begin{tabular}{c}
\textbf{Jaehyeon Park}\affilmark{1}
\quad
\textbf{Inhyeok Jeong}\affilmark{2}
\quad
\textbf{Noseong Park}\affilmark{2}\footnotemark[1]
\\[0.7em]
\affilmark{1}SolverX,
\quad
\affilmark{2}KAIST
\\[0.3em]
\texttt{jaehyeon.park@solverx.ai, inhyeok.jeong@kaist.ac.kr, noseong@kaist.ac.kr}
\end{tabular}%
}

\author{%
  Jaehyeon Park\\
  SolverX\\
  \texttt{jaehyeon.park@solverx.ai} \\
  \And
  Inhyeok Jeong \\
  KAIST \\
  \texttt{inhyeok.jeong@kaist.ac.kr} \\
  \And
  Noseong Park\footnotemark[1] \\
  KAIST \\
  \texttt{noseong@kaist.ac.kr} \\
}

\begin{document}

\maketitle
\footnotetext[1]{Corresponding Author}

\begin{abstract}
  Partial differential equation (PDE) surrogate modeling often relies on either dense observations or access to governing equations/physics constraints. In practice, observations can be sparse and partial, and the underlying physics may be partially specified or unavailable. We introduce SBNO, an equation-free paired‑data diffusion bridge matching framework that learns conditional operators for both forward and inverse PDE problems under partial observations. SBNO uses diffusion Schrödinger bridge matching to learn a pair of time-dependent drifts from a mask-aware endpoint coupling that preserves empirical input-solution pairing during training and supports both PDE-task directions within a single framework. At inference time, SBNO produces a deterministic point reconstruction by integrating the probability-flow ODE constructed from the learned drift pair, without inference-time observation or PDE guidance-weight tuning. On five PDE benchmarks with 5\% observations, SBNO achieves superior or comparable accuracy to baselines while demonstrating lower test-instance error dispersion. Furthermore, SBNO is at least $9.7\times$ faster and reduces GPU memory by at least $4.3\times$ compared to guided-diffusion baselines.
\end{abstract}

\begin{figure*}[ht]
    \centering
    \includegraphics[width=0.49\linewidth]{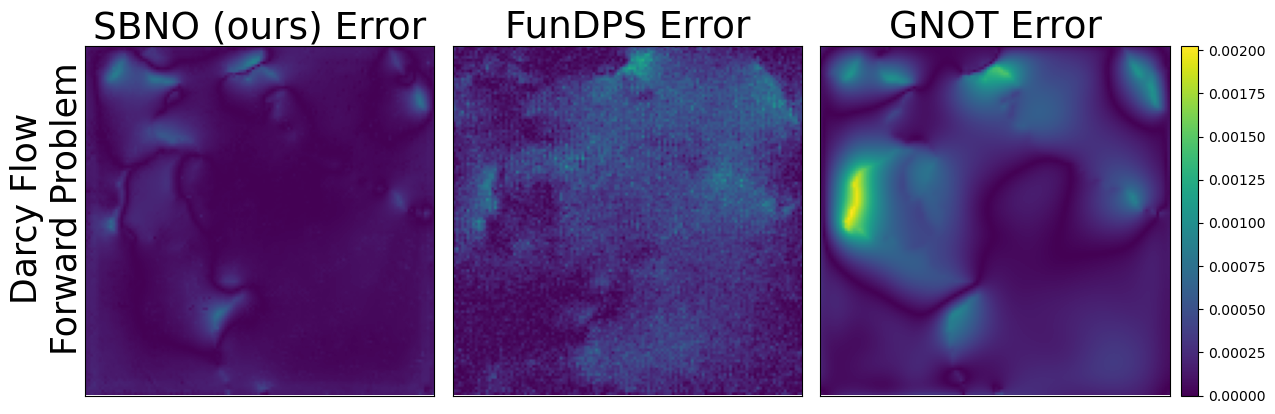}
    \includegraphics[width=0.49\linewidth]{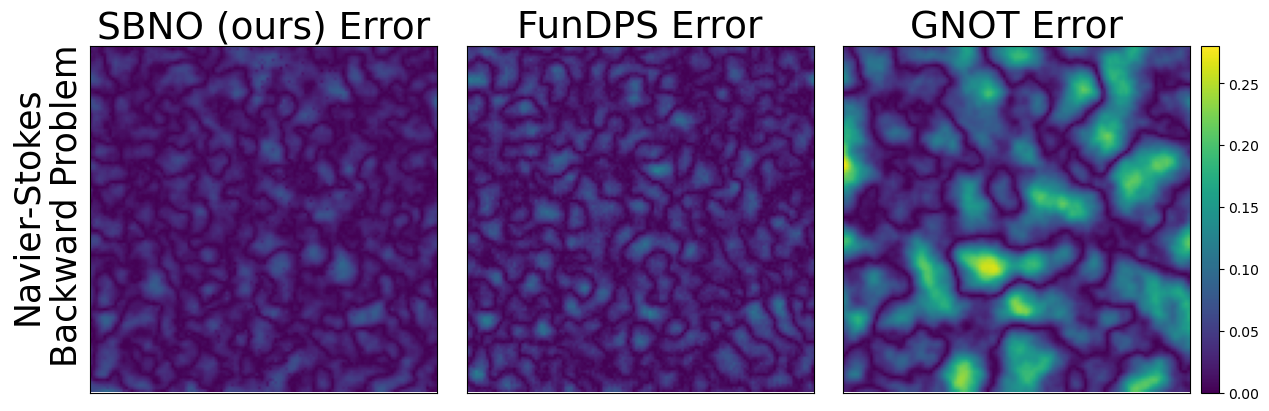}
    \caption{\textbf{Reconstruction error visualization}. We compare our method against FunDPS, a guided-diffusion baseline, and GNOT, a deterministic baseline. All error maps are visualized using a shared color scale for fair comparison, where lower values indicate better reconstruction accuracy.}
    \label{fig:teaser}
\end{figure*}

\section{Introduction} \label{sec:introduction}

Partial differential equations (PDEs) are the core mathematical language for describing complex physical phenomena in a wide range of domains, including porous media flow, wave propagation, and fluid dynamics. 
In many scientific and engineering applications, two complementary tasks frequently arise. The forward problem involves predicting the solution field given an input field such as coefficients, sources, or initial/boundary conditions. Conversely, the inverse problem aims to infer the input field from observations of the solution. This task is often ill-posed, as it may have multiple plausible solutions even when the forward model is well-defined~\citep{tarantola2005inverse, isakov2006inverse}. Although classical numerical solvers remain the gold standard for solving PDEs due to their high accuracy~\citep{quarteroni2008numerical}, their computational cost can become prohibitive in scenarios requiring PDEs to be solved repeatedly, such as design optimization, control, and inverse modeling. This has led to growing interest in learning-based surrogate models in Scientific Machine Learning (SciML)~\citep{willard2022integrating,subramanian2023towards}, which amortize computation by learning mappings from input fields to solution fields directly from data~\citep{raissi2019physics,lu2021learning,lifourier}. Despite their promise, a key limitation is their reliance on relatively complete observations during both training and evaluation. However, observations are often sparse in practice. Under these partial observation settings, the performance of conventional neural operators, such as Fourier Neural Operator (FNO)~\citep{lifourier} and Deep Operator Network (DeepONet)~\citep{lu2021learning}, can degrade significantly~\citep{huang2024diffusionpde}.

Crucially, real-world scientific and engineering settings rarely provide such idealized information. Incompleteness typically stems from two main sources. First, the governing equations may be only partially specified or entirely unknown, which limits the applicability of equation-based constraints and calibrated simulators~\citep{chien2012manufacturing, rouf2021stock, nicolaou2023data}. Second, observations are often sparse and partial due to limitations in sensor availability and incomplete domain coverage, leaving large portions of the field unobserved~\citep{myers2000improving}. These two challenges frequently co-occur, making reliable PDE-based reasoning fundamentally difficult. 
Recent generative PDE solvers are able to address partial observations using guided diffusion~\citep{dhariwal2021diffusion}, where observation consistency is enforced through a guidance term applied only at observed points~\citep{huang2024diffusionpde,yao2025guided}. While these approaches have demonstrated promise in conditioning PDE solutions on partial observations, efficient operator learning with reliable reconstruction performance in this regime remains relatively underexplored. 

To enable efficient and accurate operator learning under partial observations, we leverage diffusion bridge matching inspired by Schr\"odinger bridges and entropy-regularized transport~\citep{leonard2014reciprocal,de2021diffusion,shi2023diffusion}. 
Unlike the classical Schr\"odinger bridge problem, which optimizes an endpoint coupling under prescribed endpoint marginals, our method keeps the empirical endpoint coupling induced by paired PDE data fixed and learns the associated bridge dynamics.
This bridge-based perspective provides a smooth diffusion path between paired endpoints, while in our method the endpoint coupling itself is prescribed by the training data rather than optimized~\citep{ghosal2022stability}.
In contrast to guided diffusion-based models, our method explicitly conditions on the observation pattern by providing the network with both the masked input field and its corresponding binary mask. 
This mask-aware conditioning distinguishes observed values from zero-filled missing entries and provides explicit observation-support information for forward and inverse reconstruction. In our experiments, we additionally report the dispersion of reconstruction errors across test instances, measured by the standard deviation of relative $L_2$ errors over the test set.

Moreover, our approach offers practical advantages for deployment. In contrast to guided diffusion-based solvers, which require inference-time guidance-weight tuning as shown in Appendix~\ref{app:hyperparameter_sensitivity}, our method performs deterministic inference via the probability-flow ODE constructed from the learned drift pair~\citep{shi2023diffusion}. This enables generation of a single high-fidelity reconstruction without reliance on stochastic sampling. This leads to significantly reduced memory usage and faster inference relative to existing diffusion-based PDE solvers~\citep{huang2024diffusionpde,yao2025guided}. Finally, our approach is equation-free, requiring no access to the governing PDE during training or inference. Our method assumes access to fully observed input-solution pairs during training, but requires neither the governing PDE nor physics-based constraints during training or inference. We summarize the main contributions of this work as follows:

\begin{itemize}
\item Our method achieves better or comparable reconstruction accuracy to prior baselines while exhibiting lower error dispersion across test samples.

\item Compared to diffusion-based PDE solvers, our approach is at least $9.7\times$ faster and uses at least $4.3\times$ less GPU memory at inference time.

\item Our inference is tuning-free with no inference-time hyperparameter search and equation-free, requiring no access to the governing PDE during training and inference.
\end{itemize}


\section{Related Works} \label{sec:Related_works}

\paragraph{PDE Inverse Problems}
PDE inverse problems aim to infer unknown coefficients, sources, or boundary conditions from observations of the state and are commonly formulated as PDE-constrained optimization problems or within a Bayesian framework~\citep{stuart2010inverse, isakov2006inverse, harrach2021introduction}. While classical approaches can be accurate, they are often expensive at scale. Learning-based methods have therefore been explored to amortize inference, including PINN-style formulations~\citep{yang2021b, yu2022gradient, jagtap2022physics} and operator learning approaches~\citep{long2025invertible, molinaro2023neural, chophysics, nelsen2025operator, moss2025fnope}. 
Recent work~\citep{lingsch2024fuse} has also explored unified formulations that jointly address forward and inverse problems within a single framework. Nevertheless, under sparse or irregular observations, the mapping from observations to parameters is frequently ill-posed. These limitations motivate conditional generative formulations that can represent uncertainty and produce diverse solutions consistent with the same partial observations.


\paragraph{Function-space Generative Methods}
Function-space generative models address a different need than standard neural solvers. Instead of learning a single best prediction, they aim to represent a distribution over fields in an infinite-dimensional space.
The Generative Adversarial Neural Operator (GANO)~\citep{rahmangenerative} adapts GAN~\citep{goodfellow2020generative}-style objectives to functions by combining coordinate-based decoders with global latent variables, while the Denoising Diffusion Operator~\citep{lim2025score} uses operator parameterizations to model stochasticity more explicitly and improve generalization on grids. 
More recently, diffusion-based formulations~\citep{huang2024diffusionpde, yao2025guided} have become popular because they offer stable training and flexible conditioning and have been explored for generating PDE solutions and imputing missing regions under partial observations. 
Despite their strong empirical performance, many of these methods rely on stochastic sampling at inference or on hand-designed guidance mechanisms to enforce consistency with sparse observations, which can be costly or sensitive to conditioning design. This motivates bridge-based alternatives that incorporate partial observations into the endpoint construction and learn stochastic dynamics between the resulting field representations.


\paragraph{Diffusion Bridge and Schr\"odinger Bridge}
Schr\"odinger bridges enforce endpoint marginals by construction while staying close to a reference diffusion in KL divergence, providing a principled link to entropy-regularized optimal transport~\citep{follmer1988random, leonard2013survey}.
Building on this view, recent works have proposed scalable solvers by learning the bridge dynamics from data. \citet{de2021diffusion} introduced a Diffusion Schr\"odinger Bridge method that alternates between forward/backward processes via Iterative Proportional Fitting. Subsequent work~\citep{shi2023diffusion} improves stability and scalability via matching-style objectives, while Denoising Diffusion Bridge Models~\citep{zhou2023denoising} learn bridge scores for paired distribution translation as an alternative bridge-based formulation. In our setting, DSBM provides the bridge-matching machinery. Our contribution is a PDE-specific, mask-aware endpoint coupling and its unified use for forward and inverse reconstruction.


\section{Preliminary} \label{sec:Preliminary}

\subsection{Path Measures and a Reference Diffusion} \label{subsec:preliminary_Path_measure}

Let $\mathbf{X}_{0:T}:=(\mathbf{X}_t)_{t\in[0,T]}$ be a continuous-time stochastic process with values in $\mathbb{R}^d$. Given a set of probability measures $\mathcal{P}(\Omega)$ on path space $\Omega := C([0,T],\mathbb{R}^d)$, a \emph{path measure} $\mathbb{P} \in \mathcal{P}(\Omega)$ denotes the probability law of the entire trajectory $\mathbf{X}_{0:T}$. Its marginal distribution at time $t$ is denoted by $\mathbb{P}_t$, i.e., $\mathbb{P}_t$ is the law of $\mathbf{X}_t$ under $\mathbb{P}$. A distinguished reference path measure $\mathbb{Q}$ is fixed as the law induced by the diffusion SDE
\begin{equation} 
d\mathbf{X}_t = f_t(\mathbf{X}_t)\,dt + \sigma_t\,d\mathbf{B}_t, \qquad \mathbf{X}_0 \sim \pi_0,
\label{eq:ref_sde}
\end{equation}
where $\mathbf{B}_t$ is a standard Brownian motion, $f_t:[0,T]\times\mathbb{R}^d\to\mathbb{R}^d$ is a drift, and $\sigma_t$ is a scalar diffusion coefficient. Standard regularity assumptions, such as local Lipschitzness of $f_t$, ensure the well-posedness of~\eqref{eq:ref_sde} and the existence of the associated path measure $\mathbb{Q}$.

Given two path measures $\mathbb{P}$ and $\mathbb{Q}$ with $\mathbb{P}$ absolutely continuous with respect to $\mathbb{Q}$, the path-space Kullback-Leibler (KL) divergence is defined as:
\begin{equation}
    \mathrm{KL}(\mathbb{P}\|\mathbb{Q}) := \begin{cases} \displaystyle \mathbb{E}_{\mathbb{P}}\left[\log\frac{d\mathbb{P}}{d\mathbb{Q}}(\mathbf{X}_{0:T})\right] & \text{if } \mathbb{P} \ll \mathbb{Q}, \ \\ +\infty & \text{otherwise.} \end{cases}
\label{eq:path_kl}
\end{equation}
Crucially, if $\mathbb{P} \ll \mathbb{Q}$ is induced by an SDE with the same diffusion coefficient $\sigma_t > 0$ but a different drift $u_t(x)$, Girsanov's theorem~\citep{leonard2012girsanov} under standard Novikov-type conditions allows the KL divergence to be expressed as the expected kinetic energy of the drift difference:
\begin{equation}
\mathrm{KL}(\mathbb{P}\|\mathbb{Q}) = \frac{1}{2} \mathbb{E}_{\mathbb{P}} \left[ \int_0^T \| ( u_t - f_t)(\mathbf{X}_t) \|^2/\sigma_t^2 dt \right].
\label{eq:kl_girsanov}
\end{equation}
This formulation connects the probabilistic divergence minimization to a stochastic optimal control problem.

\subsection{Schr\"odinger Bridge and Entropy-regularized Optimal Transport} \label{subsec:preliminary_SB_EOT}
The \emph{Schr\"odinger Bridge} (SB) problem seeks to find a path measure $\mathbb{P}^*$ that minimizes the KL divergence subject to fixed initial and terminal marginal constraints $\pi_0$ and $\pi_T$ as
\begin{equation}
\mathbb{P}^* = \underset{\mathbb{P} \in \mathcal{P}(\Omega)}{\arg\min} \ \mathrm{KL}(\mathbb{P}\|\mathbb{Q}) \quad \text{s.t.} \quad \mathbb{P}_0 = \pi_0, \ \mathbb{P}_T = \pi_T.
\label{eq:sb_dynamic}
\end{equation}
Whenever $\mathbb{P}\ll \mathbb{Q}$ and both measures share the same diffusion coefficient $\sigma_t$, the Girsanov representation~\eqref{eq:kl_girsanov} connects Eq.~\eqref{eq:sb_dynamic} to a stochastic optimal control formulation. 
The Schr\"odinger bridge problem can also be interpreted as an \emph{entropic} variant of optimal transport. Among all path measures satisfying the endpoint constraints, the bridge selects the one that deviates the least from the reference dynamics $\mathbb{Q}$. Equivalently, it is the maximum-entropy solution with respect to $\mathbb{Q}$ under the same constraints. This connection becomes explicit at the level of the endpoint coupling. Let $\mathbb{Q}_{0,T}$ denote the joint law of $(X_0,X_T)$ under the reference diffusion $\mathbb{Q}$, with joint density $q_{0,T}(x,y)=\pi_0(x)\,p(0,x;T,y)$ when $X_0\sim\pi_0$ and $p(0,\cdot \ ;T,\cdot\ )$ is the transition density. Then the optimal endpoint coupling induced by the Schr\"odinger bridge solves the \emph{static} problem
\begin{equation*}
\Pi_{0,T}^\star
=\arg\min_{\Pi\in\Pi(\pi_0,\pi_T)} \mathrm{KL}\!\left(\Pi \,\middle\|\, \mathbb{Q}_{0,T}\right),
\end{equation*}
where $\Pi(\pi_0,\pi_T)$ denotes the set of couplings with marginals $\pi_0$ and $\pi_T$. Moreover, expanding the KL shows an entropy-cost trade-off:
\begin{equation*}
\mathrm{KL}\left(\Pi\|\mathbb{Q}_{0,T}\right) = \mathbb{E}_{\Pi}\left[-\log q_{0,T}(X_0,X_T)\right] - \mathcal{H}(\Pi) + \text{const},
\end{equation*}
so minimizing KL is equivalent to minimizing a transport cost induced by the reference dynamics while maximizing the coupling entropy $\mathcal{H}(\Pi)$. For instance, when $\mathbb{Q}$ is a Brownian diffusion with constant noise scale, $-\log q_{0,T}(x,y)$ reduces to a quadratic cost, and the noise intensity controls the strength of the entropic regularization. As the noise decreases, the solution approaches classical OT, whereas larger noise yields higher-entropy, more mass-splitting couplings.

When the reference $\mathbb{Q}$ is Markov with transition density $p(s,\cdot \ ;t,\cdot\ )$, assuming densities exist, the SB solution follows strictly positive potentials $\{\phi_t,\psi_t\}_{t\in[0,T]}$ such that
\begin{equation}\label{eq:positive_potential}
p_t^\star(x)=\phi_t(x)\psi_t(x)\,q_t(x),
\end{equation}
and the optimal endpoint coupling can be written as
\begin{equation}\label{eq:optimal_endpoint_coupling}
\Pi_{0,T}^\star(dx,dy)=\phi_0(x)\,p(0,x;T,y)\,\psi_T(y)\,dx\,dy,
\end{equation}
where $q_t$ denotes the marginal density of $\mathbb{Q}$ at time $t$, and $(\phi_0,\psi_T)$ are determined by the marginal constraints. It is often convenient to express a path measure through endpoint conditioning of the reference process.
Let $\mathbb{Q}_{\mid 0,T}(\cdot\mid x_0,x_T)$ denote the $\mathbb{Q}$-bridge, i.e., the conditional law of $X_{0:T}$ under $\mathbb{Q}$ given
$(X_0,X_T)=(x_0,x_T)$. Then any \emph{mixture} of $\mathbb{Q}$-bridges can be written as
\begin{equation}
\Pi(\cdot)=\int \mathbb{Q}_{\mid 0,T}(\cdot\mid x_0,x_T)\,\Pi_{0,T}(dx_0,dx_T).
\end{equation}
The endpoint-coupling view in Eq.~\eqref{eq:positive_potential}--\eqref{eq:optimal_endpoint_coupling} provides the key perspective used in our method. We will construct PDE-specific empirical endpoint couplings and learn bridge dynamics for forward and inverse reconstruction under these fixed couplings. For completeness, we defer the Markovian/reciprocal projection viewpoint of Schr\"odinger bridges and a brief review of diffusion Schr\"odinger bridge matching to Appendix~\ref{app:Theoretical_background}.


\begin{figure*}
    \centering
    \includegraphics[width=0.9\linewidth]{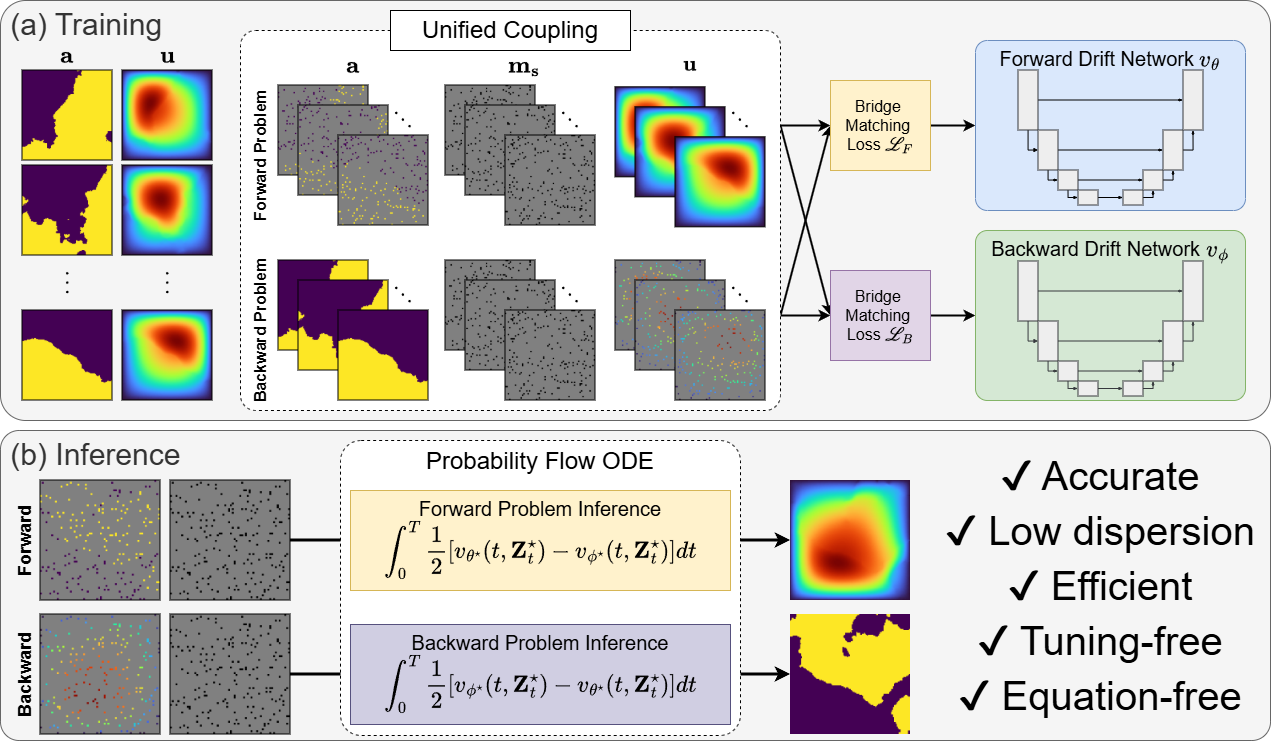}
    \caption{\textbf{Overview of SBNO. (a) Training.} From paired PDE samples $(\mathbf{a},\mathbf{u})$ and a random observation mask $\mathbf{m}_{\mathbf{s}}$, SBNO lifts each partially observed field by concatenating masked values and the binary mask. It constructs a unified endpoint coupling that mixes forward and inverse pairs while preserving empirical input-solution pairing. Forward/backward drift networks are trained with diffusion Schr\"odinger bridge matching losses. \textbf{(b) Inference.} Given partial observations and their mask, SBNO produces a deterministic reconstruction of the target field using the probability-flow ODE constructed from the learned drift pair, with the same trained model supporting both PDE tasks without observation or PDE guidance-weight tuning.}
    \label{fig:Model}
\end{figure*}

\section{Method} \label{sec:Method}
In this section, we present a model for forward and inverse PDE reconstruction under partial observations using fixed-coupling diffusion bridge matching. We begin by defining the problem and then explain how the Schr\"odinger bridge framework applies in this context. Applied to PDE data, this approach learns bridge dynamics from endpoint couplings constructed using empirical input-solution pairings and partial-observation masks.

\subsection{Problem Definition} \label{subsec:method_Problem_Definition}
Let $\mathcal{X}\subset\mathbb{R}^{d_x}$ be a spatial domain and let $\mathcal{T}\subset\mathbb{R}$ be a time domain. To cover both static and time-dependent PDEs under a unified notation, define the evaluation domain
\begin{equation}
\mathcal{Z} := \begin{cases}
\mathcal{X}, & \text{(static PDEs, e.g., Darcy flow)}\\
\mathcal{T}\times\mathcal{X}, & \text{(time-dependent PDEs).}
\end{cases}
\end{equation}
Consider a family of PDEs parameterized by an input function $\mathbf{a}\in\mathcal{A}$, where $\mathbf{a}$ may encode coefficients, source terms, or initial/boundary conditions. The corresponding solution operator is denoted by $\mathbf{u} = \mathcal{G}(\mathbf{a}) \ \text{with} \ \mathcal{G}:\mathcal{A}\to\mathcal{U},$
where $\mathcal{U}$ is a function space over $\mathcal{Z}$. The solution field is denoted by $\mathbf{u}(z)$, where $\mathbf{u}:\mathcal{Z}\to\mathbb{R}^{d_u}$ may be scalar- or vector-valued (e.g., $\mathbf{u}(\cdot,\cdot)$ for the time-dependent case and $\mathbf{u}(\cdot)$ for the static case).

Let $\mathcal{S}$ denote the space of specifications, such as sensor locations, masks, or sampling patterns. For each $\mathbf{s} \in \mathcal{S}$, we denote the corresponding partial observation of the input function as $\mathbf{a}_\mathbf{s}$, and the partial observation of the solution as $\mathbf{u}_\mathbf{s}$. 
We assume a data distribution supported on paired samples $(\mathbf{a}, \mathbf{u})$ satisfying $\mathbf{u}=\mathcal{G}(\mathbf{a})$. Under specification $\mathbf{s}\in\mathcal{S}$, the ideal forward and inverse reconstruction targets can be viewed as conditional laws:
\begin{equation}
\begin{aligned}
p_\mathrm{F}(\mathbf{u} | \mathbf{a}_\mathbf{s}, \mathbf{s}) \quad \text{consistent with} \quad \mathbf{u}=\mathcal{G}(\mathbf{a}) \quad , \quad
p_\mathrm{I}(\mathbf{a} | \mathbf{u}_\mathbf{s}, \mathbf{s}) \quad \text{consistent with} \quad \mathbf{u}=\mathcal{G}(\mathbf{a}).
\label{eq:operator_distribution}
\end{aligned}
\end{equation}
These distributions represent the probabilistic uncertainty over the unobserved regions of the solution $\mathbf{u}$ and the input parameters $\mathbf{a}$, respectively. In general, the inverse mapping from $\mathbf{u}$ to $\mathbf{a}$ is ill-posed and may be non-unique, especially under partial observations. Therefore, $p_\mathrm{I}(\mathbf{a}| \mathbf{u}_\mathbf{s},\mathbf{s})$ describes a potentially non-unique reconstruction target, for which we learn bridge dynamics and evaluate deterministic point reconstruction in the main experiments. We propose to learn bridge dynamics for these two reconstruction directions within a single unified fixed-coupling framework.

\subsection{State Representation and Bridge Endpoints} \label{subsec:method_endpoint_constructions}
To accommodate practical scenarios, we represent the fields in the function spaces $\mathcal{A}$ and $\mathcal{U}$ using discretized forms. 
After discretizing the evaluation domain $\mathcal{Z}$ with $N$ grid points, each field is stored as a tensor (e.g., of shape $(C,H,W)$ on a 2D grid) and flattened into a vector in $\mathbb{R}^D$ with total dimension $D := C\cdot N$.
Given a specification $\mathbf{s}\in\mathcal{S}$ and a discretized field $\mathbf{x}\in\mathbb{R}^{D}$, let $\mathbf{m}_{\mathbf{s}}\in\{0,1\}^{D}$ be the associated binary mask. We define the masked observation $\mathbf{x}_{\mathbf{s}} := \mathbf{m}_{\mathbf{s}}\odot \mathbf{x} \in \mathbb{R}^{D}$ and the corresponding \emph{lifted} state representation:
\begin{equation}
\mathcal{E}_{\mathbf{s}}(\mathbf{x}) :=
\begin{bmatrix} \mathbf{m}_{\mathbf{s}}\odot \mathbf{x} \\ \mathbf{m}_{\mathbf{s}} \end{bmatrix}
\in \mathbb{R}^{2D}.
\label{eq:lifted_state}
\end{equation}
For a fully observed field, the mask is taken as the all-ones vector $\mathbf{1}$, yielding $\mathcal{E}_{\mathbf{1}}(\mathbf{x})=[\mathbf{x};\mathbf{1}]$. Based on this representation, we construct the endpoint pairs for the forward and inverse problems. Let $(\mathbf{a},\mathbf{u})$ denote an input-solution pair consistent with the PDE family, i.e., $\mathbf{u}=\mathcal{G}(\mathbf{a})$. We define two distinct endpoint couplings.

For the \textbf{forward operator}, given partial input $\mathbf{s}\in\mathcal{S}$, we define the source state $\mathbf{X}_0^{\mathrm{F}} := \mathcal{E}_{\mathbf{s}}(\mathbf{a})$ and the target state $\mathbf{X}_T^{\mathrm{F}} := \mathcal{E}_{\mathbf{1}}(\mathbf{u})$ (full solution). The induced endpoint coupling is the joint law:
\begin{equation}
\Pi_{0,T}^{\mathrm{F}} := \mathrm{Law}(\mathbf{X}_0^{\mathrm{F}},\mathbf{X}_T^{\mathrm{F}})\in \mathcal{P}(\mathbb{R}^{2D}\times\mathbb{R}^{2D}),
\label{eq:coupling_forward}
\end{equation}
where $\mathrm{Law}(\cdot)$ denotes the distribution induced by the underlying PDE dataset.
Conversely, for the \textbf{inverse operator}, we define the source state $\mathbf{X}_0^{\mathrm{I}} := \mathcal{E}_{\mathbf{1}}(\mathbf{a})$ (full input) and the target state $\mathbf{X}_T^{\mathrm{I}} := \mathcal{E}_{\mathbf{s}}(\mathbf{u})$ (partial solution), with the corresponding coupling:
\begin{equation}
\Pi_{0,T}^{\mathrm{I}} := \mathrm{Law}(\mathbf{X}_0^{\mathrm{I}},\mathbf{X}_T^{\mathrm{I}}) \in \mathcal{P}(\mathbb{R}^{2D}\times\mathbb{R}^{2D}).
\label{eq:coupling_inverse}
\end{equation}

The conditional operators of interest are given by the endpoint conditionals derived from couplings:
\begin{equation}
\begin{aligned}
p_F(\cdot| x_0) := \Pi^F_{0,T}(dx_T | \mathbf{X}_0^F=x_0) \quad , \quad
p_I(\cdot| x_T) := \Pi^I_{0,T}(dx_0 | \mathbf{X}_T^I=x_T).
\label{eq:endpoint_conditionals}
\end{aligned}
\end{equation}
These endpoint conditionals define the target reconstruction laws associated with the task-specific couplings. Their exact recovery by the learned Markovian dynamics or deterministic ODE is not guaranteed. Specifically, for the forward problems, sampling $\mathbf{X}_T \sim p_\mathrm{F}(\cdot | \mathbf{X}_0^{\mathrm{F}})$ predicts the full solution $\widehat{\mathbf{u}}$ from partial input observations. For the inverse problems, sampling $\mathbf{X}_0 \sim p_\mathrm{I}(\cdot | \mathbf{X}_T^{\mathrm{I}})$ recovers the full input $\widehat{\mathbf{a}}$ from partial solution observations. The lifted representation~\eqref{eq:lifted_state} places both partial observations and full fields in the shared space $\mathbb{R}^{2D}$, enabling unified learning of the drift fields.

\subsection{Schr\"odinger-Bridge-Inspired Fixed-Coupling Formulations} \label{subsec:method_bridge_formulation}

This section formulates the forward and inverse reconstruction problems using a fixed-coupling diffusion bridge construction inspired by Schr\"odinger bridges. 
Rather than solving the classical Schr\"odinger bridge problem by optimizing over all endpoint couplings with prescribed marginals, we preserve the empirical endpoint couplings induced by paired PDE samples and learn Markovian drift fields associated with the resulting reciprocal bridge mixtures on the common state space \( \mathbb{R}^{2D} \). 

Let $\Omega := C([0,T],\mathbb{R}^{2D})$ be the path space of continuous trajectories $\mathbf{X}_{0:T}=(\mathbf{X}_t)_{t\in[0,T]}$. Fix a reference path measure $\mathbb{Q}$ on $\Omega$ induced by the SDE, $d\mathbf{X}_t = f_t(\mathbf{X}_t)\,dt + \sigma_t\,d\mathbf{B}_t,$
where $\mathbf{B}_t$ is a standard Brownian motion in $\mathbb{R}^{2D}$, $f_t:\mathbb{R}^{2D}\to\mathbb{R}^{2D}$ is a drift, and $\sigma_t>0$ is a scalar diffusion coefficient. The parameter $t\in[0,T]$ denotes the time, which is independent of the physical coordinates $z \in \mathcal{Z}$. 

Classical Schr\"odinger bridges are specified by endpoint marginal constraints and implicitly determine an entropy-regularized optimal endpoint coupling. 
In contrast, our setting provides paired PDE samples, and therefore an empirical endpoint coupling $\Pi^\alpha_{0,T}$ is already available. 
We keep this empirical endpoint coupling fixed and define a reciprocal bridge mixture $\widetilde P^\alpha := \Pi^\alpha_{0,T} Q_{|0,T},$ where $Q_{|0,T}$ denotes the reference bridge conditioned on the endpoints. 
The learning problem is then to approximate the Markovian projection of this fixed-coupling reciprocal mixture by time-dependent drift fields. 
This fixed-coupling construction preserves the empirical input-solution pairing in the training data while providing bridge dynamics that can be used for forward and inverse reconstruction.

\subsection{Unified Bridge Matching and Training Objective for Forward and Inverse Operators} \label{subsec:method_unified_BM}

To model both the forward and inverse problems within a single framework, we construct a fixed empirical endpoint coupling that mixes the two task-specific couplings while preserving the empirical input-solution pairing between \(a\) and \(u\), i.e., \(u=G(a)\). We then train a pair of time-dependent drifts on bridge samples drawn from this fixed-coupling mixture.

Two endpoint couplings $\Pi_{0,T}^{\mathrm{F}}$ and $\Pi_{0,T}^{\mathrm{I}}$ defined in Section~\ref{subsec:method_endpoint_constructions} can be combined into a single coupling by mixing these two distributions $\Pi_{0,T} \;:=\; \frac{1}{2}\,\Pi_{0,T}^{\mathrm{F}} + \frac{1}{2}\,\Pi_{0,T}^{\mathrm{I}}.$ This mixture defines the fixed reciprocal bridge mixture used for training:
\begin{equation}
\widetilde P^\Pi := \Pi_{0,T}Q_{|0,T} = \int Q_{|0,T}(\cdot\mid x_0,x_T)\, \Pi_{0,T}(dx_0,dx_T).
\end{equation}
Importantly, $\Pi_{0,T}$ is prescribed by the empirical paired-data construction and is not optimized over $\Pi(\pi_0,\pi_T)$.
Let $M\in\{\mathrm{F},\mathrm{I}\}$ be a binary indicator with $p(M=\mathrm{F})=1/2$. Sampling $(\mathbf{a},\mathbf{u})$ from the paired PDE dataset and $\mathbf{s} \sim p(\mathbf{s})$, we can define the joint random variables as:
\begin{equation}
(\mathbf{X}_0,\mathbf{X}_T) = \begin{cases}
(\mathcal{E}_{\mathbf{s}}(\mathbf{a}),\ \mathcal{E}_{\mathbf{1}}(\mathbf{u})), & M=\mathrm{F},\\
(\mathcal{E}_{\mathbf{1}}(\mathbf{a}),\ \mathcal{E}_{\mathbf{s}}(\mathbf{u})), & M=\mathrm{I},
\end{cases}
\label{eq:unified_coupling_indicator}
\end{equation}
where $\Pi_{0,T} = \mathrm{Law}(X_0,X_T)$. By construction, $\Pi_{0,T}$ is supported only on endpoint pairs generated from paired input-solution samples $(a,u)$. The mixture changes the reconstruction direction, but not the underlying input-solution pairing.

Given an endpoint pair $(\mathbf{X}_0,\mathbf{X}_T)\sim \Pi_{0,T}$ and a bridge time $\tau\sim\mathrm{Unif}[0,T]$, an intermediate state is sampled from the $\tau$-marginal of the bridge $\mathbf{X}_\tau \sim \mathbb{Q}_{|0,T}(\mathbf{X}_\tau \in \cdot | \mathbf{X}_0,\mathbf{X}_T).$
This construction yields tuples $(\mathbf{X}_0,\mathbf{X}_\tau,\mathbf{X}_T)$ drawn from the mixture-of-bridges path measure. As in DSBM, two time-dependent drift fields $v_\theta$ and $v_\phi$ on $\mathbb{R}^{2D}$ are introduced:
\begin{equation}
v_\theta:[0,T]\times\mathbb{R}^{2D}\to\mathbb{R}^{2D}, \quad v_\phi:[0,T]\times\mathbb{R}^{2D}\to\mathbb{R}^{2D}.
\label{eq:drift_param_unified}
\end{equation}
Here, $v_\theta$ approximates the forward drift and $v_\phi$ approximates the backward drift. The networks use the same architecture with separate parameters and are optimized independently.
Following the bridge-matching formulation reviewed in Section~\ref{subsec:preliminary_DSBM}, we train forward/backward drifts by regression on $\mathbb{Q}$-bridge samples. 
Define the $Q$-bridge conditional-score targets
\begin{equation}
\begin{aligned}
s^{T|t}(X_T;x) := \sigma_t^2\nabla_x \log \mathbb{Q}_{T | t}(\mathbf{X}_T | \mathbf{X}_t) \quad , \quad
s^{t|0}(x;X_0) := \sigma_t^2\nabla_x \log \mathbb{Q}_{t | 0}(\mathbf{X}_t | \mathbf{X}_0).
\end{aligned}
\end{equation}
We minimize the following regression losses
\begin{equation}
\begin{aligned}
\mathcal{L}_{\mathrm{F}}(\theta) &:=  \mathbb{E} \Bigl[ \bigl\| s^{T|\tau}(X_T;X_\tau)  - v_\theta(\tau,\mathbf{X}_\tau) \bigr\|^2 / \sigma_\tau^{2}\Bigr] , \\
\mathcal{L}_{\mathrm{B}}(\phi) &:=  \mathbb{E}\Bigl[\bigl\| s^{\tau|0}(X_\tau;X_0)  - v_\phi(\tau,\mathbf{X}_\tau) \bigr\|^2 /\sigma_\tau^{2} \Bigr], 
\end{aligned}
\label{eq:unified_loss_noalpha}
\end{equation}
where the expectation is taken over $\tau\sim\mathrm{Unif}[0,T]$, $(X_0,X_T)\sim\Pi_{0,T}$, and $X_\tau\sim \mathbb{Q}_{|0,T}(\cdot\mid X_0,X_T)$.
Because $\Pi_{0,T}$ mixes forward and inverse endpoint pairs, a single model $(v_\theta,v_\phi)$ is trained to support both partial-to-full directions through the shared lifted state space~\eqref{eq:lifted_state}. Both drifts are required for our probability-flow ODE inference, including when training uses only one PDE-task coupling. We therefore use Schr\"odinger bridge matching as a bridge-dynamics learning tool under a fixed empirical endpoint coupling, rather than as a procedure for estimating the optimal coupling between endpoint marginals.

The learned bridge induces two conditional generative processes on $\mathbb{R}^{2D}$. For the forward problems, given a partial input $\mathbf{X}_0=\mathcal{E}_{\mathbf{s}}(\mathbf{a})$, $\mathbf{X}_T$ can be sampled using the forward SDE with drift $v_\theta$. The decoded value of $\mathbf{X}_T$ yields the full solution prediction $\widehat{\mathbf{u}}$. For the inverse problems, given a partial solution $\mathbf{X}_T=\mathcal{E}_{\mathbf{s}}(\mathbf{u})$,  $\mathbf{X}_0$ can be sampled using the backward SDE with drift $v_\phi$. The decoded value of $\mathbf{X}_0$ yields the full input prediction $\widehat{\mathbf{a}}$. The detailed training procedure can be found in Appendix~\ref{app:training_algorithm}.

\subsection{Deterministic Inference via Probability-flow ODE} \label{subsec:method_deterministic_inference}
Although inference can be performed stochastically by sampling trajectories from the learned SDEs, we adopt a deterministic approach to obtain a stable representative reconstruction. Specifically, we solve the associated \emph{probability flow ODEs} of the learned bridge. 
For the fixed-coupling bridge mixture \(\widetilde P^\Pi\), let $P^\Pi := \mathrm{proj}_{\mathcal M}(\widetilde P^\Pi)$ denote its Markovian projection, with one-time marginal density $p_t^\Pi$. The corresponding forward and backward Markovian projection drifts, denoted by $v_F^\Pi$ and $v_B^\Pi$, satisfy the standard forward-backward drift relation:
\begin{equation}
v_B^\Pi(t,x) = -v_F^\Pi(t,x) + \sigma_t^2 \nabla \log p_t^\Pi(x).
\label{eq:dsbm_vphi_vtheta_relation}
\end{equation}
Under the standard regularity assumptions and the corresponding marginal initialization, this identity yields a probability-flow ODE with the same one-time marginals as the Markovian projection $P^\Pi$
Concretely, the probability flow $\mathbf{Z}_t^\Pi$ has the same one-time marginals as $\mathbb{P}^\Pi$, i.e.,
$\mathrm{Law}(\mathbf{Z}_t^\Pi)=\mathbb{P}_t^\Pi$ for all \(t\in[0,T]\), and is governed by:
\begin{equation}
dZ_t^\Pi = \left\{ f_t(Z_t^\Pi) + \frac{1}{2} \left[ v_F^\Pi(t,Z_t^\Pi) - v_B^\Pi(t,Z_t^\Pi) \right] \right\}dt.
\label{eq:probability_flow_ode}
\end{equation}
In practice, ($v_F^\Pi$ ,$v_B^\Pi$) are replaced by the learned networks ($v_\theta$, $v_\phi$). For forward inference we initialize $\mathbf{Z}_0 = \mathbf{X}_0^\mathrm{F} = \mathcal{E}_\mathbf{s}(\mathbf{a})$ and for inverse inference we initialize $\mathbf{Z}_T = \mathbf{X}_T^\mathrm{I}=\mathcal{E}_\mathbf{s}(\mathbf{u})$.
Note that this probability-flow ODE matches only the one-time marginals $\mathbb{P}_t^\Pi$ of the Markovian projected fixed-coupling bridge. 
It does not reproduce the full stochastic path law of the bridge mixture, nor does it preserve the empirical endpoint coupling $\Pi_{0,T}$. 
Consequently, the deterministic endpoint relation induced by the ODE generally differs from the fixed empirical coupling used during bridge matching. For a fixed observation, its endpoint is a deterministic point reconstruction, not necessarily the conditional mean, and it cannot represent multimodal uncertainty. However, for PDE solving where the goal is a representative reconstruction rather than sampling diverse solutions, this deterministic flow provides a practical single prediction following the learned marginal evolution.


\section{Experiments} \label{sec:experiments}
In this section, we present experiments designed to validate the proposed method in terms of PDE reconstruction accuracy, error dispersion, and inference efficiency. They are conducted under noiseless observations, and SBNO inference is performed deterministically throughout. This setting allows for a clearer comparison of the three aspects above under a controlled evaluation protocol. To further examine broader use cases, we provide additional experiments on noisy partial observations and stochastic inference for predictive-uncertainty diagnostics in Appendix~\ref{app:noisy_observation} and Appendix~\ref{app:uncertainty_quantification}, respectively.

\subsection{Experimental Setup} \label{subsec:experimental_setup}

\paragraph{Datasets}
We evaluate both forward and inverse problem-solving tasks under partial observations on five PDE benchmarks: four datasets from the DiffusionPDE~\cite{huang2024diffusionpde} and one dataset from The Well~\citep{ohana2024well}. All datasets are discretized on a $(128 \times 128)$ spatial grid. During inference, we utilize only 5\% of the available observations, corresponding to 819 points. Details about the datasets can be found in Appendix~\ref{app:datasets}.
Beyond the five main benchmarks, we further evaluate SBNO on three smaller-scale PDE datasets with distinct dynamical characteristics, as reported in Appendix~\ref{app:additional_experiments}.

\begin{table}[ht]
\vspace{-1mm}
\centering
\caption{Qualitative comparison of SBNO and the baseline configurations evaluated in this study.}
\label{table:qualitative_comparison}
\resizebox{\textwidth}{!}{%
\begin{tabular}{lccccc}
\toprule
Capability & SBNO (ours) & FunDPS \& DiffusionPDE & FNOPE & GNOT & FNO \\
\midrule
Observation conditioning & mask-aware & only observation & only observation & only observation & mask-aware \\
Unified PDE-task training & \textbf{yes} & \textbf{yes} & no & no & no \\
Inference-time hyperparameters & \textbf{not required} & required & \textbf{not required} & \textbf{not required} & \textbf{not required} \\
PDE information & \textbf{equation-free} & PDE guidance & \textbf{equation-free} & \textbf{equation-free} & \textbf{equation-free} \\
Inference type & \textbf{both possible} & stochastic & stochastic & deterministic & deterministic \\
\bottomrule
\end{tabular}
}\vspace{-1mm}
\end{table}

\paragraph{Baseline Methods}
We compare SBNO against representative baselines from three categories: guided diffusion-based PDE solvers, deterministic neural operators, and probabilistic neural operators. 
Specifically, we include DiffusionPDE~\citep{huang2024diffusionpde} and FunDPS~\citep{yao2025guided} as diffusion-based baselines, FNO~\citep{lifourier} and GNOT~\citep{hao2023gnot} as deterministic neural-operator baselines, and FNOPE~\citep{moss2025fnope} as a probabilistic operator-learning baseline. 
For methods that do not natively share a unified forward/inverse objective, we train separate models for each task when applicable. 
Further details on the baseline input representations, inference protocols, and implementation choices are provided in Appendix~\ref{app:baselines}. 
We summarize the qualitative comparison between SBNO and the baselines in Table~\ref{table:qualitative_comparison}. The FunDPS results are obtained from our reproduced configurations; we did not recover its originally reported accuracy, and Appendix E documents the protocol and unresolved reproduction discrepancy.

\paragraph{Training and Inference Protocol}
We parameterize the forward/backward drifts of SBNO, $v_\theta$ and $v_\phi$, using a time-conditioned U-Net. At inference time, SBNO produces reconstructions deterministically by integrating the probability-flow ODE induced by the learned Schr\"odinger bridge. We solve this ODE using a fourth-order Runge-Kutta (RK4) solver with 32 integration steps. Accordingly, during training, we compute the bridge-matching objective by sampling only the discrete time points that coincide with the 32-step RK4 trajectory, and we resample sensor locations at every iteration while keeping the observation ratio fixed at 5\% in the main experiments. For diffusion-based baselines, we tune inference-time hyperparameters on the validation set. A hyperparameter sensitivity analysis is provided in Appendix~\ref{app:hyperparameter_sensitivity}, and the exact configurations for each model are reported in Appendix~\ref{app:hyperparameter}.

\subsection{Results} \label{subsec:results}
\paragraph{Prediction Error}
We evaluate reconstruction accuracy using the relative $L_2$ error and report both the mean and the test-instance error dispersion, namely ``std,'' as described in Appendix~\ref{app:Evaluation}. Table~\ref{table:main_results} summarizes results on the benchmark datasets under 5\% partial observations. Overall, SBNO achieves the best or second-best mean relative $L_2$ error in all ten forward and inverse benchmark settings. Moreover, SBNO attains relatively low dispersion among competitive baselines with comparable mean performance, reducing the average error dispersion by 44.91\%, from $0.0416$ for FunDPS to $0.0229$. 
These results suggest that SBNO combines strong average accuracy with lower test-instance error dispersion in the sparse-observation regime. For all the considered baselines, SBNO remains competitive against diffusion-based methods while also outperforming the deterministic and probabilistic operator baselines in most settings.

\begin{table*}[t]
\captionsetup{skip=1.5pt}
\caption{The evaluation results are reported on five benchmark datasets. Each performance metric is expressed as the ``mean $\pm$ std'' of the relative $L_2$ error across test samples. For clarity, the best values in each column are highlighted in \textbf{bold}, while the second-best values are marked with \underline{underline}.}
\label{table:main_results}
\small
\centering
\resizebox{1.0\textwidth}{!}{%
\begin{tabular}{llcccccc}
\toprule
\textbf{Dataset} & \textbf{Problem} & \textbf{SBNO (Ours)} & \textbf{FunDPS} & \textbf{DiffusionPDE} & \textbf{FNOPE} & \textbf{GNOT} & \textbf{FNO} \\
\midrule
\multirow{2}{*}{Darcy flow} & Forward & \textbf{0.0359} $\pm$ \textbf{0.0133}    & \underline{0.0461} $\pm$ 0.0207             &  0.0552 $\pm$ 0.0849                          &  0.2219 $\pm$ 0.0893 & 0.0510  $\pm$ 0.0148  &  0.0582 $\pm$ \underline{0.0135}\\
                            & Inverse & \textbf{0.0269} $\pm$ \textbf{0.0125}    & 0.0835 $\pm$ 0.0380                         &  \underline{0.0499} $\pm$ \underline{0.0338}  & 0.2241 $\pm$ 0.1110 & 0.0779  $\pm$ 0.0367  & 0.0946 $\pm$ 0.0409   \\ 
\midrule
\multirow{2}{*}{Poisson}    & Forward & \textbf{0.0424} $\pm$ \textbf{0.0219}    & \underline{0.0775} $\pm$ 0.0473             &  0.0986 $\pm$ 0.1696                          &  0.7849 $\pm$ 0.4061 &  0.0929 $\pm$ \underline{0.0428}  &  0.1458 $\pm$ 0.0518  \\
                            & Inverse & \underline{0.2167} $\pm$ \textbf{0.0345} & 0.3664 $\pm$ 0.1232                         &  \textbf{0.2015} $\pm$ \underline{0.0792}     &  0.8423 $\pm$ 0.2244 &  0.3585 $\pm$ 0.0910  &  0.3912 $\pm$ 0.0991  \\
\midrule
\multirow{2}{*}{Helmholtz}  & Forward & \textbf{0.0519} $\pm$ \underline{0.0315} & \underline{0.0563} $\pm$ \textbf{0.0310}    &  0.0902 $\pm$ 0.1412                          &  0.7388 $\pm$ 0.3684 &  0.0926 $\pm$ 0.0393  &  0.2191 $\pm$ 0.0854 \\
                            & Inverse & \underline{0.2310} $\pm$ \textbf{0.0501} & 0.2550 $\pm$ \underline{0.0740}             &  \textbf{0.1893} $\pm$ 0.2557                 &  0.7111 $\pm$ 0.2000 &  0.3728 $\pm$ 0.0865  &  0.4178 $\pm$ 0.0947 \\
\midrule
\multirow{2}{*}{\shortstack[l]{Nonbounded\\Navier-Stokes}}             
                            & Forward & \textbf{0.0368} $\pm$ \textbf{0.0086}    & \underline{0.0435} $\pm$ \underline{0.0136} &  0.0922 $\pm$ 0.2189                          &  1.1290 $\pm$ 0.2625 &  0.2360 $\pm$ 0.0476  &  0.2470 $\pm$ 0.0437 \\
                            & Inverse & \textbf{0.1097} $\pm$ \textbf{0.0167}    & \underline{0.1241} $\pm$ \underline{0.0234} &  0.1584 $\pm$ 0.2362                          &  1.8253 $\pm$ 0.4081 &  0.3669 $\pm$ 0.0629  &  0.4106 $\pm$ 0.0557 \\
\midrule
\multirow{2}{*}{Gray-Scott} & Forward & \textbf{0.0393} $\pm$ \underline{0.0193}    &  0.0846 $\pm$ 0.0222            &  \underline{0.0468} $\pm$ 0.0545              &  0.0930 $\pm$ \textbf{0.0109} &  0.1025 $\pm$ 0.0084  &  0.0857 $\pm$ 0.0112  \\
                            & Inverse & \textbf{0.0382} $\pm$ \underline{0.0167}    &  0.0683 $\pm$ 0.0225            &  \underline{0.0589} $\pm$ 0.0622              &  0.1094 $\pm$ \textbf{0.0089} &  0.0871 $\pm$ 0.0084  &  0.0724 $\pm$ 0.0102 \\
\bottomrule
\end{tabular}%
}
\vspace{-3mm}
\end{table*}

\begin{wraptable}{r}{0.4\textwidth}
\centering
\caption{Cost comparison between diffusion-based models. We report wall-clock time per sample and peak GPU memory. The best values are highlighted in \textbf{bold}.}
\label{table:inference_efficiency}
\resizebox{0.4\textwidth}{!}{%
\begin{tabular}{lcc}
\toprule
Method & \shortstack[c]{Time per\\Sample (sec)} & \shortstack[c]{Peak Mem\\(MB)} \\
\midrule
DiffusionPDE & 46.65         & 3,846.7       \\
FunDPS       & 2.71          & 5,428.8       \\
SBNO (ours)  & \textbf{0.28} & \textbf{886.0} \\
\bottomrule
\end{tabular}
}
\end{wraptable}

\paragraph{Inference Efficiency}
We next evaluate the inference-time efficiency of SBNO under the same partial-observation setting. This comparison is restricted to the diffusion-based baselines, FunDPS and DiffusionPDE, as they are the most relevant generative baselines for comparison with SBNO. We evaluate both metrics over the test set, as described in Appendix~\ref{app:Evaluation}. All results are averaged over five runs. Table~\ref{table:inference_efficiency} summarizes the findings. Overall, SBNO demonstrates substantially lower inference cost than the guided diffusion baselines. It is at least $9.7\times$ faster in wall-clock time per sample and consumes at least $4.3\times$ less peak GPU memory. Combined with its competitive reconstruction accuracy, these measurements indicate a favorable accuracy–cost trade-off relative to the reproduced guided-diffusion baselines in the evaluated setting.
We additionally analyze the trade-off between reconstruction performance and inference cost by varying the ODE solver steps, with results reported in Appendix~\ref{app:ODE_solver_step}.

\section{Conclusion} \label{sec:conclusion}
In this work, we address the problem of PDE forward and inverse operator learning under sparse observations. 
We introduce SBNO, a Schr\"odinger-bridge-inspired neural operator that uses a mask-aware PDE endpoint coupling and diffusion bridge matching to support forward and inverse reconstruction with one trained drift pair. It performs deterministic point reconstruction through the probability-flow ODE constructed from the learned drifts, without observation or PDE guidance-weight tuning. Across five PDE benchmarks with 5\% point-wise observations, SBNO achieves competitive or superior reconstruction accuracy while reducing test-instance error dispersion. Moreover, SBNO significantly reduces inference cost compared to guided diffusion solvers, being at least $9.7\times$ faster and using at least $4.3\times$ less peak GPU memory, making it a practical and scalable alternative for repeated reconstructions under sparse observations.

\paragraph{Limitations and Future Works}
While the results are strong in the current benchmark setting, the present implementation of SBNO still assumes discretized fields defined on a grid. As a result, although the model can handle partial observations on that grid through masking, it cannot directly ingest off-grid observations at arbitrary continuous sensor coordinates without first projecting or interpolating them onto the underlying grid. For fixed observations, deterministic ODE inference returns a single point reconstruction and does not guarantee recovery of the conditional mean or preservation of the empirical endpoint coupling, nor can it represent multimodal uncertainty. This limitation stems from the current grid-based backbone rather than from the Schr\"odinger bridge formulation itself. Extending SBNO to architectures that can natively process irregular sensor observations is an important direction for future work.


\clearpage



\bibliography{SBNO.bib}
\bibliographystyle{plainnat}

\clearpage


\appendix

\section{Additional Theoretical Background} \label{app:Theoretical_background}
\subsection{Markovian and Reciprocal Projections} \label{subsec:preliminary_Markov}
The Schr\"odinger bridge $\mathbb{P}^\star$ can be characterized through two structural constraints: (i) Markovianity and (ii) preservation of the $\mathbb{Q}$-bridge. ``Preserving the $Q$-bridge'' means that the conditional path law given endpoints 
$P(\cdot\mid X_0=x_0,X_T=x_T)=Q_{|0,T}(\cdot\mid x_0,x_T)$ is fixed for all $(x_0,x_T)$, while only the endpoint coupling may vary.
These are formalized via two projection operators acting on path measures. Under mild assumptions, the $\mathbb{Q}$-bridge is itself a diffusion bridge and satisfies the Doob $h$-transform representation
\begin{equation}
\begin{aligned}
d\mathbf{X}^{0,T}_t = &\Bigl\{f_t(\mathbf{X}^{0,T}_t) +\sigma_t^2 \nabla \log \mathbb{Q}_{T|t}(x_T |\mathbf{X}^{0,T}_t) \Bigr\}dt \\& + \sigma_t\, d\mathbf{B}_t,
\quad \mathbf{X}^{0,T}_0=x_0,
\label{eq:q_bridge_sde}
\end{aligned}
\end{equation}
where $\mathbb{Q}_{T| t}(\cdot| x)$ denotes the conditional density of $\mathbf{X}_T$ given $\mathbf{X}_t=x$ under $\mathbb{Q}$ and $\nabla$ denotes the gradient with respect to the state variable $X_t$. 

\paragraph{Markovian Projection}
Consider a path measure $\Pi \in \mathcal{P}(\Omega)$ (typically a mixture of bridges). Let $\mathcal{M}$ be the class of Markov diffusion processes with diffusion coefficient $\sigma_t$. The \emph{Markovian projection} of $\Pi$, denoted $\mathbb{M}^\star=\mathrm{proj}_{\mathcal{M}}(\Pi)$, is the unique element in $\mathcal{M}$ under standard conditions. It is associated with the SDE:
\begin{equation}
d\mathbf{X}^\star_t = \bigl\{ f_t(\mathbf{X}^\star_t) + v_t^\star(\mathbf{X}^\star_t) \bigr\}dt + \sigma_t d\mathbf{B}t,
\label{eq:markovian_projection_sde}
\end{equation}
where the drift adjustment $v_t^\star$ is given by the conditional expectation over the target mixture $\Pi$:
\begin{equation}
v_t^\star(x) = \sigma_t^2 \mathbb{E}_{\Pi}\!\left[ \nabla \log \mathbb{Q}_{T|t}(\mathbf{X}_T|x) \big| \mathbf{X}_t = x \right].
\label{eq:markovian_projection_vstar}
\end{equation}
Crucially, $\mathbb{M}^\star$ minimizes the reverse KL divergence from the mixture $\Pi$ to the Markov manifold $\mathcal{M}$~\citep{shi2023diffusion}:
\begin{equation}
\mathbb{M}^\star = \underset{\mathbb{M}\in\mathcal{M}}{\arg\min} \ \mathrm{KL}(\Pi\|\mathbb{M}).
\label{eq:markovian_projection_reverseKL}
\end{equation}
This projection effectively "marginalizes" the non-Markovian history dependencies into the current state via the drift term.

\paragraph{Reciprocal Projection}
The \emph{reciprocal class} $\mathcal{R}(\mathbb{Q})$ consists of all measures that share the same bridge dynamics as $\mathbb{Q}$ but may have arbitrary endpoint couplings~\citep{leonard2014reciprocal}. Formally,
\begin{equation}
\begin{aligned}
\mathcal{R}(\mathbb{Q}) := \bigl\{\Pi\in\mathcal{P}(\Omega) : \exists\Pi_{0,T}\ \text{s.t.}\ \Pi=\Pi_{0,T}\mathbb{Q}_{| 0,T} \bigr\}.
\label{eq:reciprocal_class}
\end{aligned}
\end{equation}
The \emph{reciprocal projection} of a path measure $\mathbb{P}$ onto $\mathcal{R}(\mathbb{Q})$ simply replaces the bridge of $\mathbb{P}$ with that of $\mathbb{Q}$ while preserving the endpoint joint distribution~\citep{shi2023diffusion}:
\begin{equation}
\mathrm{proj}_{\mathcal{R}(\mathbb{Q})}(\mathbb{P}) := \mathbb{P}_{0,T}\mathbb{Q}_{| 0,T}.
\label{eq:reciprocal_projection}
\end{equation}

\paragraph{SB as an Intersection}
The reciprocal projection ensures the correct bridge dynamics but may break Markovianity. Conversely, the Markovian projection ensures Markovianity but may distort the bridge. Among path measures satisfying $P_0=\pi_0$ and $P_T=\pi_T$, the Schr\"odinger bridge is the unique element in $\mathcal{M}\cap\mathcal{R}(Q)$. 

\subsection{Diffusion Schr\"odinger Bridge Matching} \label{subsec:preliminary_DSBM}
\citet{shi2023diffusion} proposed \emph{Iterative Markovian Fitting} (IMF) to compute the Schr\"odinger bridge by alternating projections between the Markov manifold $\mathcal{M}$ and the reciprocal class $\mathcal{R}(Q)$. Starting from an initial reciprocal path measure $P^{(0)}\in\mathcal{R}(Q)$ that satisfies the endpoint constraints $\mathbb{P}^{(0)}_0=\pi_0$ and $\mathbb{P}^{(0)}_T=\pi_T$, IMF defines the sequence $\{\mathbb{P}^{(n)}\}_{n\in\mathbb{N}}$ by
\begin{equation}
\mathbb{P}^{(2n+1)}=\mathrm{proj}_{\mathcal{M}}\!\big(\mathbb{P}^{(2n)}\big),\
\mathbb{P}^{(2n+2)}=\mathrm{proj}_{\mathcal{R}(Q)}\!\big(\mathbb{P}^{(2n+1)}\big),
\label{eq:imf}
\end{equation}
where both projections preserve the endpoint marginals for all $n$. Under suitable conditions, the fixed point of this alternating-projection scheme coincides with the Schr\"odinger bridge $P^\star$. 
Since Markovian projection $\mathrm{proj}_{\mathcal{M}}$ is generally intractable in high dimensions, DSBM~\citep{shi2023diffusion} provides a scalable implementation by replacing the exact Markovian projection with regression-based \emph{bridge matching}. Concretely, given a Markov iterate $\mathbb{M}^{(n)}\in\mathcal{M}$, the reciprocal update forms a reciprocal measure by preserving the current endpoint coupling:
\begin{equation}
\widetilde{\mathbb{P}}^{(n)} := \mathrm{proj}_{\mathcal{R}(\mathbb{Q})}\big(\mathbb{M}^{(n)}\big)
= \mathbb{M}^{(n)}_{0,T}\mathbb{Q}_{|0,T},
\label{eq:rec_update}
\end{equation}
i.e., a mixture of $Q$-bridges with endpoint coupling $\mathbb{M}^{(n)}_{0,T}=\mathrm{Law}_{\mathbb{M}^{(n)}}(X_0,X_T)$. DSBM then updates the forward/backward drift fields by regressing to conditional-score targets of the reference $Q$-bridge evaluated on bridge samples, producing the next Markov iterate $M^{(n+1)}\approx \mathrm{proj}_{\mathcal{M}}(\widetilde{P}^{(n)})$. The updated endpoint coupling $\mathbb{M}^{(n+1)}_{0,T}$ is subsequently reused in Eq.~\eqref{eq:rec_update} for the next reciprocal update.

We defer the explicit regression losses and the sampling procedure to Section~\ref{subsec:method_unified_BM}, where we instantiate them with our PDE-specific endpoint constructions and the unified coupling.

\section{Training Algorithm} \label{app:training_algorithm}

\begin{algorithm*}[ht]
\caption{Unified DSBM bridge matching training for forward and inverse operators}
\label{alg:unified_dsbm_training}
\begin{algorithmic}[1]
\REQUIRE Paired PDE samples $(\mathbf{a},\mathbf{u})$ with $\mathbf{u}=\mathcal{G}(\mathbf{a})$, specification distribution $p(\mathbf{s})$ and lifted map $\mathcal{E}_{\mathbf{s}}(\cdot)$ in Eq.~\eqref{eq:lifted_state},  reference diffusion $\mathbb{Q}$ with bridge sampler $\mathbb{Q}_{| 0,T}(\cdot| \mathbf{X}_0,\mathbf{X}_T)$ and conditional score oracles $\nabla\log \mathbb{Q}_{T| t}(\mathbf{X}_T| \mathbf{X}_t)$, $\nabla\log \mathbb{Q}_{t| 0}(\mathbf{X}_t| \mathbf{X}_0)$, diffusion schedule $\sigma_t$, time horizon $T$, batch size $B$, iteration number $N$, time-conditioned U-Net drift networks $v_\theta(t,\mathbf{x})$ and $v_\phi(t,\mathbf{x})$.
\FOR{$n=1,\dots,N$}
    \STATE Sample a batch $\{(\mathbf{a}^{(i)},\mathbf{u}^{(i)})\}_{i=1}^B$,  specifications $\{\mathbf{s}^{(i)}\}_{i=1}^B \sim p(\mathbf{s})$, and mode indicators $M^{(n)}$ with $p(M^{(n)}=\mathrm{F})=1/2$.
    \IF{$M^{(i)}=\mathrm{F}$}
        \STATE $\mathbf{X}_0^{(i)}, \mathbf{X}_T^{(i)} \gets \mathcal{E}_{\mathbf{s}^{(i)}}(\mathbf{a}^{(i)}), \mathcal{E}_{\mathbf{1}}(\mathbf{u}^{(i)})$ \rcomment{Forward: partial input $\rightarrow$ full solution}
    \ELSE 
        \STATE $\mathbf{X}_0^{(i)}, \mathbf{X}_T^{(i)} \gets \mathcal{E}_{\mathbf{1}}(\mathbf{a}^{(i)}), \mathcal{E}_{\mathbf{s}^{(i)}}(\mathbf{u}^{(i)})$
        \rcomment{Inverse: partial solution $\rightarrow$ full input}
    \ENDIF
    \STATE Sample $\tau^{(i)} \sim \mathrm{Unif}[0,T]$, intermediate state $\mathbf{X}_{\tau^{(i)}}^{(i)} \sim \mathbb{Q}_{|0,T}(\mathbf{X}_{\tau^{(i)}}^{(i)} \in \cdot | \mathbf{X}_0^{(i)},\mathbf{X}_T^{(i)}).$
    \STATE Compute minibatch losses (Monte Carlo estimates of Eq.~\eqref{eq:unified_loss_noalpha})\\
    $\widehat{\mathcal{L}}_{\mathrm{F}}(\theta) \gets \frac{1}{B}\sum_{i=1}^B \frac{1}{\sigma_{\tau^{(i)}}^{2}} \left\| \sigma_{\tau^{(i)}}^{2}\nabla \log \mathbb{Q}_{T| \tau^{(i)}}\!\Bigl(\mathbf{X}_T^{(i)}| \mathbf{X}_{\tau^{(i)}}^{(i)}\Bigr) - v_\theta\!\bigl(\tau^{(i)},\mathbf{X}_{\tau^{(i)}}^{(i)}\bigr) \right\|^2$ \\
    $\widehat{\mathcal{L}}_{\mathrm{B}}(\phi) \gets \frac{1}{B}\sum_{i=1}^B \frac{1}{\sigma_{\tau^{(i)}}^{2}} \left\| \sigma_{\tau^{(i)}}^{2}\nabla \log \mathbb{Q}_{\tau^{(i)}| 0}\!\Bigl(\mathbf{X}_{\tau^{(i)}}^{(i)}| \mathbf{X}_0^{(i)}\Bigr) - v_\phi\!\bigl(\tau^{(i)},\mathbf{X}_{\tau^{(i)}}^{(i)}\bigr) \right\|^2.$
    \STATE Total loss: $\widehat{\mathcal{L}}(\theta,\phi)\gets \widehat{\mathcal{L}}_{\mathrm{F}}(\theta)+\widehat{\mathcal{L}}_{\mathrm{B}}(\phi).$
    \STATE Update parameters $(\theta,\phi)$ by minimizing $\widehat{\mathcal{L}}(\theta,\phi)$.
\ENDFOR
\STATE \textbf{Return} $v_\theta(t,\mathbf{x})$ and $v_\phi(t,\mathbf{x})$
\end{algorithmic}
\end{algorithm*}

\subsection{Analytic Bridge Sampling and Regression Targets}

Analytic bridge sampling and regression targets. The conditional-score targets in Algorithm 1 do not require separately trained score networks. For a zero-drift Gaussian reference
$$dX_t=\sigma_t\,dB_t,$$
with a deterministic scalar diffusion schedule, define the accumulated variance
$$V(t):=\int_0^t \sigma_s^2\,ds.$$
Its transition laws are
$$ Q_{t|0}(\cdot|X_0) = \mathcal N\!\left(X_0,V(t)I\right),
\qquad Q_{T|t}(\cdot|X_t) = \mathcal N\!\left(X_t,[V(T)-V(t)]I\right). $$
For \(0<t<T\), a state conditioned on the endpoint pair \((X_0,X_T)\) can therefore be sampled as
$$ X_t = \left(1-\frac{V(t)}{V(T)}\right)X_0 + \frac{V(t)}{V(T)}X_T + \sqrt{\frac{V(t)[V(T)-V(t)]}{V(T)}}\,\epsilon,
\qquad \epsilon\sim\mathcal N(0,I). $$
The scaled conditional-score targets in Eq. (17) are
$$ s_{T|t}(X_T;X_t) = \sigma_t^2 \frac{X_T-X_t}{V(T)-V(t)}, \qquad s_{t|0}(X_t;X_0) = \sigma_t^2 \frac{X_0-X_t}{V(t)}. $$

For constant diffusion \(\sigma_t=\sigma\) and \(T=1\), these expressions reduce to
$$ X_t=(1-t)X_0+tX_T+\sigma\sqrt{t(1-t)}\,\epsilon, \qquad s_{T|t}(X_T;X_t)=\frac{X_T-X_t}{1-t},
\qquad s_{t|0}(X_t;X_0)=\frac{X_0-X_t}{t}.
$$
The drift networks directly regress these analytic targets using the weighted objectives in Eq.~\eqref{eq:unified_loss_noalpha}. A small time cutoff is used to avoid evaluating the singular targets at the endpoints. Thus, “score oracle” denotes an analytically computable regression target, rather than an auxiliary learned model.

\section{Evaluation}
\label{app:Evaluation}

We evaluate model performance using the relative $L_2$ error between predicted and ground-truth solution fields on the test set. For each $i$-th test sample , given the true solution $y_i$ and the model prediction $\hat{y}_i$, the relative error is computed as:
\begin{equation}
e_i = \frac{| y_i - \hat{y}_i |_p}{| y_i |_p}
\end{equation}
where $| \cdot |_p$ denotes the $L_p$ norm (typically $p=2$ in our experiments). The mean and standard deviation of the relative errors over all test samples are then reported to summarize overall model accuracy and variability.
\begin{equation}
\begin{aligned}
\text{dataset error mean} = \frac{1}{N}\sum_{i=1}^{N} e_i \quad , \quad  \text{dataset error std} = \sqrt{\frac{1}{N}\sum_{i=1}^{N} \left(e_i - \frac{1}{N}\sum_{i=1}^{N} e_i \right)^2}
\end{aligned}
\end{equation}

We also report two efficiency metrics during inference: wall-clock time per sample and peak GPU memory usage. Wall-clock time is measured on a single NVIDIA B200 GPU using the largest feasible batch size for each method. To isolate pure inference cost, we exclude data loading. Peak memory is measured as the maximum GPU memory allocated during inference with a batch size of one. All results are averaged over five runs. Because time per sample is measured using method-specific batch sizes, it reflects batched inference throughput and should not be interpreted as batch-one request latency. The memory measurements use a separate batch-one protocol.


\section{Datasets} \label{app:datasets}

The main experiments in Section~\ref{sec:experiments} consist of five PDE datasets: four datasets from the DiffusionPDE benchmark release~\citep{huang2024diffusionpde} and one Gray-Scott reaction-diffusion dataset from The Well~\citep{ohana2024well}.\footnote{Data source for the DiffusionPDE benchmark release: https://github.com/jhhuangchloe/DiffusionPDE. Data source of The Well: https://huggingface.co/polymathic-ai} The four DiffusionPDE datasets include three static PDEs and one time-dependent Navier-Stokes dataset. For these four datasets, all fields are defined on the unit square domain \(\Omega=(0,1)^2\) and discretized on a \(128\times128\) grid. 
Each dataset contains 50,000 training samples, and we use fixed validation and test sets of 100 and 900 samples, respectively.

The Gray-Scott dataset differs from the DiffusionPDE datasets in both its source and construction. It is obtained from The Well and consists of two-channel reaction-diffusion trajectories. Following the construction described in Appendix~\ref{app:gray_scott}, we form one-step forecasting pairs from stored trajectory frames, yielding 10,000 training samples and 500 test samples. We include Gray-Scott in the main benchmark to evaluate SBNO on a multi-channel reaction-diffusion system with pattern-forming dynamics beyond the DiffusionPDE benchmark suite.

In addition to the five main datasets, Appendix~\ref{app:additional_experiments} reports smaller-scale experiments on three time-dependent PDE datasets from APEBench~\citep{koehler2024apebench}: Korteweg-De Vries, Kuramoto-Sivashinsky, and Kolmogorov flow. These additional datasets are used only for complementary evaluation and are not included in the main benchmark tables.

Although FunDPS originally included the bounded Navier-Stokes dataset, we exclude this dataset from our experiments due to anomalies identified in the version provided by the original authors. 
We provide further discussion and analysis of this issue in Appendix~\ref{app:flaw_dataset}.

\subsection{Time-independent Static PDEs} \label{app:static_PDE}
Following the standard dataset construction procedure used in Fourier Neural Operator (FNO) benchmarks~\citep{lifourier}, we first sample Gaussian random fields $\boldsymbol{\mu}$ on $\Omega$ and then construct the corresponding coefficient field $\mathbf{a}$. For each sample, we solve the governing equation with a second-order finite-difference scheme on the $128\times 128$ grid. Homogeneous Dirichlet boundary conditions are enforced by applying an appropriate boundary mollifier to the solution. All three static PDEs are based on Gaussian random fields $\boldsymbol{\boldsymbol{\mu}} \sim \mathcal{N}(0,(-\Delta + 9\mathbf{I})^{-2})$ defined over the spatial domain $\Omega$.

\subsubsection{Darcy Flow}
Darcy flow models the steady-state flow of an incompressible viscous fluid through a porous medium.
We construct a piecewise-constant permeability field $\mathbf{a}$ by thresholding a Gaussian random field:
$\mathbf{a}(x)=12$ if $\boldsymbol{\mu}(x)\ge 0$ and $\mathbf{a}(x)=3$ otherwise.
The solution $\mathbf{u}$ satisfies
\begin{equation}
-\nabla\cdot\big(\mathbf{a}(x)\nabla \mathbf{u}(x)\big)=f(x), \quad x\in\Omega,
\label{eq:Darcy_Flow}
\end{equation}
with homogeneous Dirichlet boundary conditions $\mathbf{u}\vert_{\partial\Omega}=0$ and $f(x)=1$.

\subsubsection{Poisson Equation}
The Poisson equation describes steady-state diffusion with a source term.
We set the coefficient field as $\mathbf{a}=\boldsymbol{\mu}$ and solve
\begin{equation}
\nabla^2\mathbf{u}(x)=\mathbf{a}(x), \quad x\in\Omega,
\label{eq:Poisson_equation}
\end{equation}
under homogeneous Dirichlet boundary conditions $\mathbf{u}\vert_{\partial\Omega}=0$.

\subsubsection{Helmholtz Equation}
The Helmholtz equation models wave propagation in heterogeneous media. We set the coefficient field as $\mathbf{a}=\boldsymbol{\mu}$ and the solution $\mathbf{u}$ satisfies
\begin{equation}
\nabla^2\mathbf{u}(x)+k^2\mathbf{u}(x)=\mathbf{a}(x), \quad x\in\Omega,
\label{eq:Helmholtz_equation}
\end{equation}
with $k=1$ and $\mathbf{u}\vert_{\partial\Omega}=0$.

\subsection{Time-dependent PDEs} \label{app:time_PDEs}
\subsubsection{Incompressible Navier-Stokes Equation}
Our time-dependent benchmarks consist of the 2D incompressible Navier-Stokes equations in an unbounded (periodic) setting on the unit torus in vorticity form. The initial vorticity field $w_0$ is sampled from a Gaussian random field on $\Omega$
\begin{equation}
w_0 \sim \mathcal{N}\!\left(0,\; 7^{1.5}(-\Delta + 49\mathbf{I})^{-2.5}\right).
\end{equation}

We consider the 2D Navier-Stokes equations for a viscous incompressible fluid on the unit torus,
written in vorticity form:
\begin{equation}
\begin{aligned}
\partial_t w(x,t) + \mathbf{u}(x,t)\cdot \nabla w(x,t) &= \nu \Delta w(x,t) + q(x),
&& x \in \Omega,\; t \in (0,T], \\
\nabla \cdot \mathbf{u}(x,t) &= 0,
&& x \in \Omega,\; t \in [0,T], \\
w(x,0) &= w_0(x),
&& x \in \Omega,
\end{aligned}
\end{equation}
where $w = \nabla \times \mathbf{u}$ denotes the vorticity and we set the viscosity to $\nu = 10^{-3}$.
The forcing term follows a fixed spatial pattern for $x=(x_1,x_2)$:
\[
q(x) = \frac{1}{10}\Big(\sin(2\pi(x_1+x_2)) + \cos(2\pi(x_1+x_2))\Big).
\]
We simulate the dynamics up to a terminal time corresponding to one second using a pseudo-spectral solver,
and store the vorticity fields on a $128\times 128$ grid.

\subsubsection{Gray-Scott Equation} \label{app:gray_scott}
The Gray-Scott system describes the spatiotemporal evolution of two interacting chemical species through coupled reaction-diffusion dynamics and is widely used as a benchmark for pattern formation. Following The Well dataset~\citep{ohana2024well}, we consider two concentration fields $A(x,t)$ and $B(x,t)$ on a two-dimensional spatial domain, governed by

\begin{equation}
\begin{aligned}
\frac{\partial A}{\partial t} = \delta_A \Delta A - AB^2 + f(1-A), \qquad
\frac{\partial B}{\partial t} = \delta_B \Delta B - AB^2 - (f+k)B,
\end{aligned}
\end{equation}

where $f$ and $k$ denote the feed and kill rates, respectively, and $\delta_A,\delta_B$ are the diffusion coefficients. In the original dataset, all simulations use fixed diffusion coefficients, the fields are defined on a uniform Cartesian grid over $[-1,1]\times[-1,1]$, and periodic boundary conditions are imposed. The data contain two channels corresponding to the concentrations of species $A$ and $B$, and the initial conditions are generated from two families: random Fourier series and random clusters of Gaussians.

In our experiments, we use the parameter setting with $f=0.029$ and $k=0.057$. Each trajectory consists of 1001 time steps on a $128\times128$ grid, with simulation time step 1 second and stored frames separated by $\Delta t=10$ seconds. From each trajectory, we construct one-step forecasting pairs by taking the full field at one stored time step as input and the full field at the next stored time step as output, yielding 100 input-output pairs per trajectory. Using trajectories generated from 100 initial conditions, we obtain 10,000 training samples in total. For evaluation, we apply the same construction to trajectories generated from 5 unseen initial conditions, resulting in 500 test samples.

\subsubsection{Korteweg-De Vries Equation}
The Korteweg--De Vries equation models nonlinear dispersive wave propagation and is a classical benchmark for studying the interplay between advection, dispersion, and dissipation in time-dependent PDEs. Following the APEBench dataset~\cite{koehler2024apebench}, we consider a scalar field $u(x,t)$ on a two-dimensional spatial domain, governed by
\begin{equation}
\frac{\partial u}{\partial t} = -\frac{b}{2}(\mathbf{1}\cdot \nabla)u +\xi \, \mathbf{1}\cdot \nabla^{3}u -\zeta \, \mathbf{1}\cdot \nabla^{4}u,
\end{equation}
where $b$, $\xi$, and $\zeta$ control the advection, dispersive, and dissipative terms, respectively. In the original dataset, the field is defined on a uniform Cartesian grid over $[-1,1]\times[-1,1]$, and periodic boundary conditions are imposed. The initial condition is generated from a truncated Fourier series. In one dimension, this takes the form
\begin{equation}
u(t=0,x)=o+\sum_{k=1}^{K} a_k \sin\!\left(k\frac{2\pi}{L}x\right)+b_k \cos\!\left(k\frac{2\pi}{L}x\right),
\end{equation}
where the offset $o$ and the Fourier coefficients are sampled uniformly, and higher-dimensional initial conditions are constructed analogously by combining modes up to cutoff $K$.

In our experiments, we use the parameter setting with $b=-2.0$, $\xi=-14.0$, and $\zeta=-9.0$. Each trajectory is simulated on a $160\times160$ grid with time step interval $\Delta t=1$. The training set consists of 50 trajectories, each with 50 time steps, while the validation and test data are drawn from 30 trajectories, each with 100 time steps. From each trajectory, we construct one-step forecasting pairs by taking the full field at a given time step as input and the full field 10 time steps later as the target. This yields 40 input-target pairs per trajectory for training, resulting in 2,000 training samples in total. For validation and test, we construct 20 input-target pairs per trajectory from the 30 trajectories, yielding 600 samples in total, of which 100 are used for validation and 500 for testing.

\subsubsection{Kuramoto-Sivashinsky Equation}
The Kuramoto-Sivashinsky equation models nonlinear spatiotemporal dynamics arising from the competition between destabilizing growth, nonlinear transport, and higher-order dissipation, and is a standard benchmark for chaotic time-dependent PDEs. Following the APEBench dataset~\cite{koehler2024apebench}, we consider a scalar field $u(x,t)$ on a two-dimensional spatial domain, governed by
\begin{equation}
\frac{\partial u}{\partial t} = -\frac{b}{2}\|\nabla u\|^2 -\nu \nabla \cdot \nabla u -\zeta \, \mathbf{1}\cdot \nabla^{4}u,
\end{equation}
where $b$, $\nu$, and $\zeta$ control the nonlinear, second-order, and fourth-order terms, respectively. In the original dataset, the field is defined on a uniform Cartesian grid over $[-1,1]\times[-1,1]$, and periodic boundary conditions are imposed. The initial condition is generated from a truncated Fourier series, the same as the Korteweg-de Vries equation.

In our experiments, we use the parameter setting with $b=-6.0$, $\nu=-1.5$, and $\zeta=-15.0$. Each trajectory is simulated on a $160\times160$ grid with a time step interval $\Delta t=1$. The training set consists of 50 trajectories, each with 50 time steps, while the validation and test data are drawn from 30 trajectories, each with 100 time steps. From each trajectory, we construct forecasting pairs by taking the full field at a given time step as input and the full field 10 time steps later as the target. This yields 40 input-target pairs per trajectory for training, resulting in 2,000 training samples in total. For validation and test, we construct 20 input-target pairs per trajectory from the 30 trajectories, yielding 600 samples in total, of which 100 are used for validation and 500 for testing.

\subsubsection{Kolmogorov Flow}
Kolmogorov flow is a standard benchmark for forced two-dimensional incompressible turbulence and is widely used to study nonlinear transport and coherent structure formation in time-dependent PDEs. Following the APEBench dataset, we consider a scalar vorticity field $u(x,t)$ on a two-dimensional spatial domain, governed by
\begin{equation}
\frac{\partial u}{\partial t} = -b \left( \begin{bmatrix} 1\\ -1 \end{bmatrix} \odot \nabla(\Delta^{-1}u) \right)\cdot \nabla u +\nu \nabla \cdot \nabla u +\lambda u -k\cos\!\left(k\frac{2\pi}{L}y\right),
\end{equation}
where $\nu$ denotes the viscosity, $\lambda$ controls the linear term, and $k$ determines the forcing wavenumber. In the original dataset, the field is defined on a uniform Cartesian grid over $[-1,1]\times[-1,1]$, and periodic boundary conditions are imposed. The initial condition is generated from a truncated Fourier series, as described for the Korteweg-de Vries equation above.

In our experiments, we use the parameter setting corresponding to Reynolds number $100$, with $\nu=0.01$, $\lambda=-0.1$, and $k=4$. As in the Korteweg-de Vries equation, each trajectory is simulated on a $160\times160$ grid with time step interval $\Delta t=1$. The training set consists of 50 trajectories, each with 50 time steps, while the validation and test data are drawn from 30 trajectories, each with 100 time steps. We construct forecasting pairs in the same way as for the Korteweg-de Vries equation, by taking the full field at a given time step as input and the full field 10 time steps later as the target. This yields 40 input-target pairs per trajectory for training, resulting in 2,000 training samples in total. For validation and test, we construct 20 input-target pairs per trajectory from the 30 trajectories, yielding 600 samples in total, of which 100 are used for validation and 500 for testing.

\subsection{Data Quality Issue in the Accessed Bounded Navier‑Stokes Release} \label{app:flaw_dataset}

\begin{figure}[ht]
    \centering
    \includegraphics[width=1\linewidth]{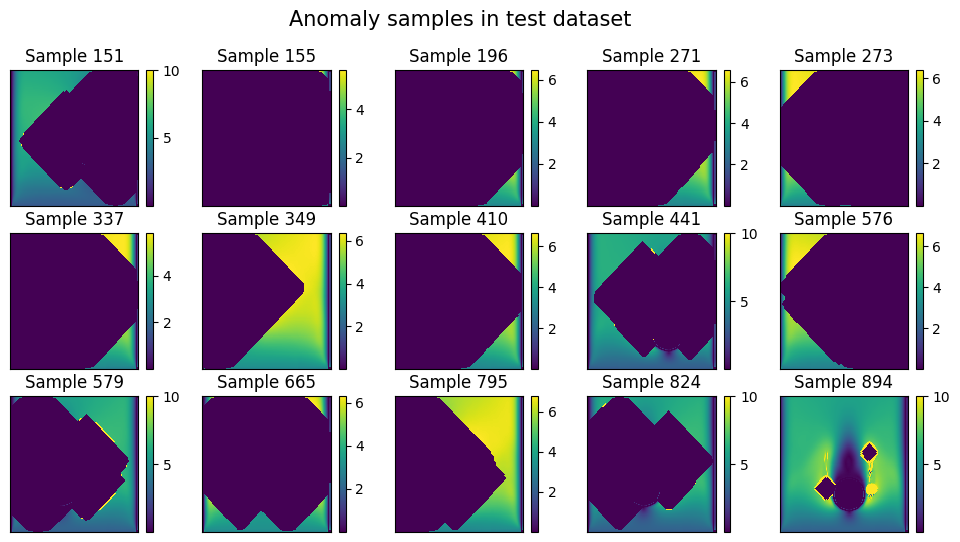}
    \caption{Visualizations of samples with data quality issue.}
    \label{fig:data_quality_issue}
\end{figure}

Upon examining the datasets provided by FunDPS, we identified data quality issues in the bounded Navier–Stokes dataset. This dataset is intended to represent 2D incompressible flow around a cylindrical obstacle in a bounded domain. However, we observed anomalies likely arising from incorrect handling of the region near the cylindrical obstacle, suggesting a flaw in the preprocessing or boundary treatment in that area. Below, we provide visualizations of the affected samples to illustrate the observed issues.


\section{Baseline Details}
\label{app:baselines}

We provide additional details on the baseline methods used in Section~\ref{sec:experiments}. 
The diffusion-based baselines are DiffusionPDE~\citep{huang2024diffusionpde} and FunDPS~\citep{yao2025guided}. 
Both methods solve reconstruction problems through guided diffusion sampling, where observation consistency is imposed during inference through guidance terms rather than by concatenating an explicit binary mask channel to the model input. 
For FunDPS, we use its multi-resolution inference strategy in the main experiments, which uses low-resolution sampling for 80\% of the inference trajectory to reduce computational cost. 
For completeness, we also evaluate a fixed-resolution variant of FunDPS in Appendix~\ref{app:detailed_results}.

We used the official FunDPS implementation and retrained its model directly at $128\times128$, following the reported architecture and training configuration for that resolution. Inference used 500 sampling steps, and only the observation and PDE guidance weights were selected using the validation split. Training was conducted on four NVIDIA B200 GPUs. Direct-resolution training should be distinguished from inference resolution: the main FunDPS results use its multi-resolution sampling strategy, whereas Appendix G.1 evaluates fixed-resolution inference.

Despite this protocol, we did not reproduce the accuracy reported in the original FunDPS paper. Our evaluation uses 5\% observations, whereas its original reported setting uses 3\%. Other differences include direct-$128\times128$ training rather than the original two-stage multi-resolution training pipeline, and a different hardware and distributed-training configuration. We cannot establish which, if any, of these differences explains the discrepancy. Consequently, comparisons with FunDPS describe our reproduced configurations and should not be interpreted as establishing superiority over its originally reported performance.

For deterministic neural-operator baselines, we include FNO~\citep{lifourier} and GNOT~\citep{hao2023gnot}. 
These two models differ in how they represent sparse observations. 
FNO operates on grid-structured inputs; therefore, for fair comparison with SBNO, we provide FNO with the same masked-grid representation consisting of the observed values and the corresponding binary mask. 
In contrast, GNOT can directly process partial observations without requiring them to be represented on the same masked grid, making it a relevant baseline for sparse-observation settings.

We also include FNOPE~\citep{moss2025fnope} as a probabilistic neural-operator baseline. 
FNOPE is particularly relevant for inverse problems because it is designed to model uncertainty in operator-learning settings. 
However, unlike SBNO, the neural-operator baselines considered here are not trained with a single unified objective that simultaneously supports both forward and inverse inference. 
Accordingly, when a baseline does not natively provide a unified forward/inverse model, we train separate models for the two tasks.

All neural-operator baselines are equation-free in the sense that they do not require access to the governing PDE or PDE residual during training or inference. 
The key distinction is that SBNO combines equation-free learning with a fixed empirical bridge coupling and a single trained pair of temporal drift networks supporting both PDE tasks, whereas the deterministic and probabilistic operator baselines are trained directly for task-specific reconstruction.


\section{Hyperparameter Sensitivity of FunDPS in Inference} 
\label{app:hyperparameter_sensitivity}

Guidance-based diffusion solvers require careful tuning of inference-time hyperparameters to achieve good reconstruction under sparse observations, in contrast to SBNO, which performs deterministic inference via the probability-flow ODE. This section reports the sensitivity trends of the guided diffusion baselines used in our experiments in Section~\ref{sec:experiments}.\footnote{We follow the same evaluation metric as Appendix~\ref{app:Evaluation}.} To evaluate hyperparameter sensitivity, we sweep the guidance weight $\zeta$ for each baseline and measure the resulting reconstruction performance under the same evaluation protocol as in the main experiments. The quantitative comparison is presented in Table~\ref{table:hp_sensitivity}.

\begin{table}[htbp] 
\captionsetup{skip=1.5pt}
\caption{Sensitivity of FunDPS to the inference-time guidance weight $\zeta$ under sparse observations. We report the relative $\ell_2$ error for both forward and inverse tasks for five PDE benchmarks. The symbol ``div'' indicates that the inference procedure fails to converge. \label{table:hp_sensitivity}}
\small
\centering
\resizebox{\linewidth}{!}{%
\begin{tabular}{lcccccccccc}
\toprule
\multirow{2}{*}{\begin{tabular}[c]{@{}c@{}}\textbf{Guidance}\\ \textbf{Weight $\zeta$}\end{tabular}}  & \multicolumn{2}{c}{\textbf{Darcy Flow}} & \multicolumn{2}{c}{\textbf{Poisson}} & \multicolumn{2}{c}{\textbf{Helmholtz}} & \multicolumn{2}{c}{\begin{tabular}[c]{@{}c@{}}\textbf{Nonbounded}\\ \textbf{Navier-Stokes}\end{tabular}} & \multicolumn{2}{c}{\textbf{Gray-Scott}} \\ 
\cmidrule(lr){2-3} \cmidrule(lr){4-5} \cmidrule(lr){6-7} \cmidrule(lr){8-9} \cmidrule(lr){10-11}
& Forward & Inverse & Forward & Inverse & Forward & Inverse & Forward & Inverse & Forward & Inverse \\
\midrule
 500    & 0.0480 & 0.2567  & 0.1649  & 0.4540  & 0.1484  & 0.2249  & 0.0624  & 0.1102   & 0.1032  & 0.0947  \\
 1,000  & 0.0374 & 0.1913  & 0.0931  & 0.3827  & 0.0725  & 0.2017  & 0.0427  & 0.1043   & 0.0920  & 0.0837  \\
 5,000  & 0.0371 & 0.1083  & 0.0775  & 0.2724  & 0.0519  & 0.2598  & 0.0358  & 0.2395   & 0.0718  & 0.0640  \\
 10,000 & 0.0449 & 0.0895  & 0.0993  & 0.2416  & 0.0499  & 0.4516  & 0.0373  & 0.5372   & 0.0662  & 0.0599  \\
 15,000 & 0.0557 & 0.0821  & 0.1324  & 0.2272  & 0.0495  & 0.7019  & 0.0426  & 0.4976   & 0.0677  & 0.0616  \\
 20,000 & 0.1254 & 0.0783  & 0.1943  & 0.2193  & 0.1201  & 0.9818  & 0.0518  & 0.3776   & 0.0695  & 0.0640  \\
 25,000 & div    & 0.0763  & div     & 0.2304  & 0.2594  & 1.2256  & 0.0648  & 0.2481   & 0.0712  & 0.0657  \\
 30,000 & div    & 0.0751  & div     & 0.2876  & 0.3199  & 1.3348  & 0.3151  & 0.2170   & div     & 0.0666  \\
 40,000 & div    & 0.0747  & div     & div     & div     & 1.2531  & 0.5091  & 0.3845   & div     & 0.1071  \\
 50,000 & div    & 0.0742  & div     & div     & div     & 1.4033  & div     & 0.4242   & div     & div     \\
\bottomrule
\end{tabular}%
}
\end{table}

\begin{figure}[ht]
    \centering
    \includegraphics[width=0.49\linewidth]{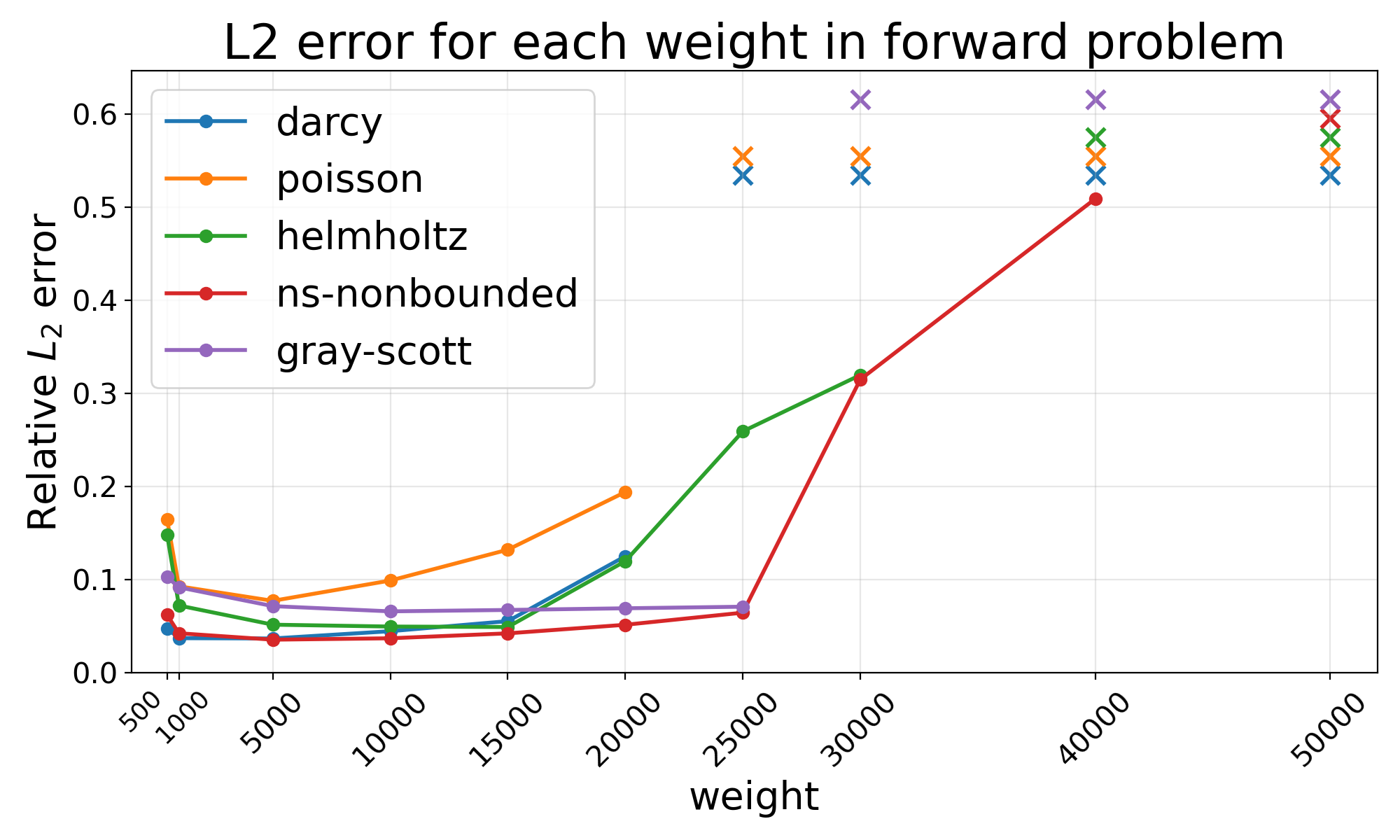}
    \includegraphics[width=0.49\linewidth]{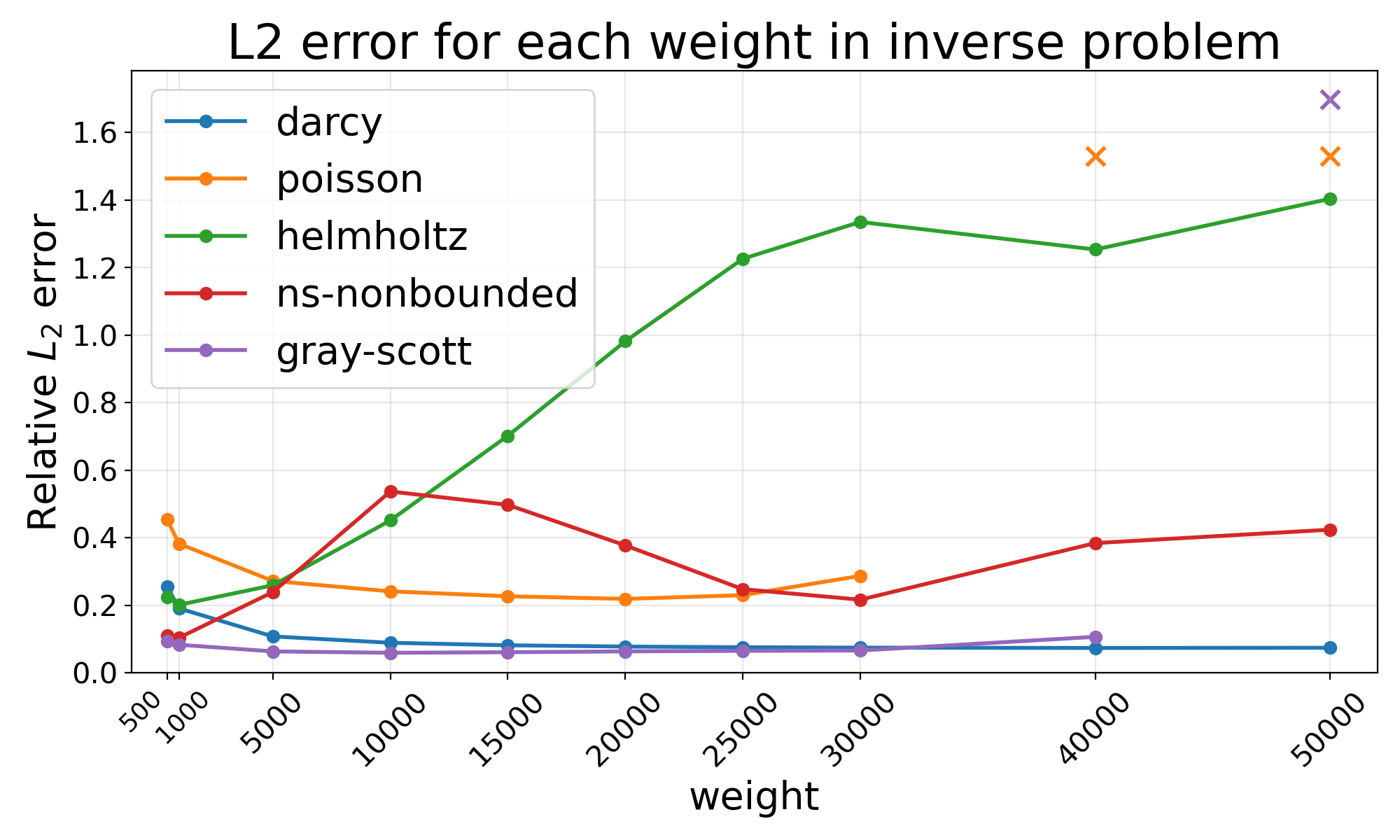}
    \caption{FunDPS sensitivity to guidance weight $\zeta$ in the forward and inverse problem. Divergent cases are marked with $\times$.}
    \label{fig:FunDPS_hp_sen}
\end{figure}

Overall, both guided diffusion baselines exhibit non-negligible sensitivity to the guidance weight, and the sensitivity depends on the dataset. For sufficiently large $\zeta$, baselines occasionally fail to converge, resulting in invalid samples or numerical divergence. We mark a run as ``div'' if it produces NaNs/Infs, diverges, or fails to return a valid reconstruction within the maximum inference budget. Figure~\ref{fig:FunDPS_hp_sen} plots the performance as a function of the guidance weight for FunDPS.
For runs that do not converge, we mark an ``$\times$'' at the corresponding guidance weight on the x-axis, using the same color as the associated curve.

Table~\ref{table:hp_sensitivity} and Figure~\ref{fig:FunDPS_hp_sen} show that FunDPS is highly sensitive to the inference-time guidance weight $\zeta$, and the degree of sensitivity varies substantially across datasets and between forward and inverse tasks. In several forward problems, performance improves as $\zeta$ increases from small values, but only up to a task-dependent range, after which the error increases again or the inference becomes unstable. This trend is visible, for example, in Darcy flow, Poisson, Helmholtz, nonbounded Navier-Stokes, and Gray-Scott, where moderate guidance generally yields the best reconstruction accuracy, while overly large guidance leads to degradation or divergence. By contrast, the inverse problems often require larger guidance weights to reach their best performance, but they also tend to be more unstable, with sensitivity becoming especially pronounced in Helmholtz and nonbounded Navier-Stokes.

Another notable trend is that the optimal $\zeta$ differs significantly across problems. Forward tasks typically favor moderate guidance, whereas inverse tasks often benefit from stronger guidance but are more prone to instability. For sufficiently large $\zeta$, inference may produce invalid samples or diverge entirely, as reflected by the “div” entries in Table~\ref{table:hp_sensitivity}. This reinforces the practical limitation of guidance-based diffusion solvers: their performance depends heavily on careful inference-time tuning, in contrast to SBNO, which uses a fixed deterministic inference procedure.

\section{Detailed Experimental Results}
\label{app:detailed_results}

\subsection{Fixed-resolution FunDPS}
As discussed in Section~\ref{subsec:experimental_setup}, we additionally evaluate FunDPS under a fixed-resolution inference setting. We denote this variant as ``FunDPS fixed'' in our results. The corresponding performance is reported in Table~\ref{table:detailed_results}, providing a direct comparison with other methods under consistent resolution settings.

\begin{table*}[ht]
\captionsetup{skip=1.5pt}
\caption{The evaluation results are reported on five benchmark datasets. Each performance metric is expressed as the ``mean $\pm$ standard deviation'' of the relative $L_2$ error across test set samples. For clarity, the best mean and best standard deviation in each column are highlighted in \textbf{bold}, while the second-best values are marked with \underline{underline}.}
\label{table:detailed_results}
\small
\centering
\resizebox{1.0\textwidth}{!}{%
\begin{tabular}{llcccccccc}
\toprule
\textbf{Dataset} & \textbf{Problem} & \textbf{SBNO (Ours)} & \textbf{FunDPS fixed} & \textbf{FunDPS} & \textbf{DiffusionPDE} & \textbf{FNOPE} & \textbf{GNOT} & \textbf{FNO}  \\
\midrule
\multirow{2}{*}{Darcy flow} & Forward &   \textbf{0.0359} $\pm$ \textbf{0.0133} & \underline{0.0442} $\pm$ 0.0248 &  0.0461 $\pm$ 0.0207&  0.0552 $\pm$ 0.0849 &   0.2219 $\pm$ 0.0893  &   0.0510 $\pm$ 0.0148 &  0.0582 $\pm$ \underline{0.0135}    \\
                            & Inverse &   \textbf{0.0269} $\pm$ \textbf{0.0125} &  0.0742 $\pm$ 0.0348 &  0.0835 $\pm$ 0.0380 &  \underline{0.0499} $\pm$ \underline{0.0338}  &   0.2241 $\pm$ 0.1110  & 0.0779 $\pm$ 0.0367   &    0.0946 $\pm$ 0.0409  \\
\midrule
\multirow{2}{*}{Poisson}    & Forward &   \textbf{0.0424} $\pm$ \textbf{0.0219} &  \underline{0.0775} $\pm$ \underline{0.0420} &  \underline{0.0775} $\pm$ 0.0473 &  0.0986 $\pm$ 0.1696  &   0.7849 $\pm$ 0.4061  &  0.0929 $\pm$ 0.0428   &  0.1458 $\pm$ 0.0518   \\
                            & Inverse &   \underline{0.2167} $\pm$ \textbf{0.0345} &  0.2193 $\pm$ 0.0609 &  0.3664 $\pm$ 0.1232 &  \textbf{0.2015} $\pm$ \underline{0.0792}  &   0.8423 $\pm$ 0.2244  & 0.3585 $\pm$ 0.0910 &  0.3912 $\pm$ 0.0991     \\
\midrule
\multirow{2}{*}{Helmholtz}  & Forward &   \underline{0.0519} $\pm$ {0.0315} &  \textbf{0.0495} $\pm$ \textbf{0.0266} &  0.0563 $\pm$ \underline{0.0310} &  0.0902 $\pm$ 0.1412   &   0.7388 $\pm$ 0.3684  &  0.0926 $\pm$ 0.0393  &  0.2191 $\pm$ 0.0854   \\
                            & Inverse &   0.2310 $\pm$ \textbf{0.0501} &  \underline{0.2017} $\pm$ \underline{0.0532} &  0.2550 $\pm$ 0.0740 &  \textbf{0.1893} $\pm$ 0.2557  &   0.7111 $\pm$ 0.2000   &  0.3728 $\pm$ 0.0865   &  0.4178 $\pm$ 0.0947  \\
\midrule
\multirow{2}{*}{\shortstack[l]{Nonbounded\\Navier-Stokes}}             
                            & Forward & \underline{0.0368} $\pm$ \textbf{0.0086} &  \textbf{0.0358} $\pm$ \underline{0.0109} &  0.0435 $\pm$ 0.0136 &  0.0922 $\pm$ 0.2189   &   1.1290 $\pm$ 0.2625  &  0.2360 $\pm$ 0.0476  &  0.2470 $\pm$ 0.0437   \\
                            & Inverse & \underline{0.1097} $\pm$ \textbf{0.0167} &  \textbf{0.1043} $\pm$ \underline{0.0187} &  0.1241 $\pm$ 0.0234 &  0.1584 $\pm$ 0.2362   &   1.8253 $\pm$ 0.4081  & 0.3669 $\pm$ 0.0629  &  0.4106 $\pm$ 0.0557 \\
\midrule
\multirow{2}{*}{Gray-Scott} & Forward & \textbf{0.0391} $\pm$ \underline{0.0212} &  0.0662 $\pm$ 0.0262 &  0.0846 $\pm$ 0.0222 &  \underline{0.0468} $\pm$ 0.0545   &   0.0930 $\pm$ \textbf{0.0109}  &  0.1025 $\pm$ 0.0084  &  0.0857 $\pm$ 0.0112  \\
                            & Inverse & \textbf{0.0364} $\pm$ \underline{0.0188} &  0.0599 $\pm$ 0.0254 &  0.0683 $\pm$ 0.0225 &  \underline{0.0589} $\pm$ 0.0622   &   0.1094 $\pm$ \textbf{0.0089}  &  0.0871 $\pm$ 0.0084  &  0.0724 $\pm$ 0.0102 \\  
\bottomrule
\end{tabular}%
}
\end{table*}

The evaluation results show that fixed-resolution inference generally yields better performance compared to the multi-resolution setting for FunDPS, with some cases even outperforming the proposed SBNO. However, this improvement in accuracy comes at a significant computational cost: the inference time per sample increases to $4.61$ seconds, which is $1.7\times$ slower than FunDPS with multi-resolution and $16.5\times$ slower than SBNO. This comparison illustrates the accuracy–cost trade-off between SBNO and the evaluated FunDPS inference configurations.

\subsection{Sensitivity to Inference Seeds}

To assess the variability of the reported evaluation metrics with respect to randomness in the inference procedure, we repeat SBNO inference with five different random seeds and report the resulting variation in Table~\ref{table:Seed_inference}. 
For each seed, we compute the relative $L_2$ error over the full test set and summarize the test-set error distribution by its mean and standard deviation. 
The table reports the average of these quantities across the five seeds together with their standard deviation across seeds.

\begin{table}[htbp] 
\captionsetup{skip=1.5pt}
\caption{Sensitivity of SBNO to inference randomness over five different seeds. For each seed, we compute the test-set mean and standard deviation of the relative $L_2$ error. Each entry reports the average and standard deviation of the corresponding metric across seeds.} \label{table:Seed_inference}
\small
\centering
\resizebox{\linewidth}{!}{%
\begin{tabular}{lcccccccc}
\toprule
\multirow{2}{*}{Problem}  & \multicolumn{2}{c}{\textbf{Forward problem}} & \multicolumn{2}{c}{\textbf{Inverse problem}} \\
\cmidrule(lr){2-3} \cmidrule(lr){4-5}
& mean & std & mean & std \\
\midrule
\textbf{Darcy Flow}  & 0.0359 $\pm$ 0.000708   & 0.0133 $\pm$ 0.000627   & 0.0269 $\pm$ 0.000297   & 0.0125 $\pm$ 0.000185 \\
\midrule
\textbf{Poisson}     & 0.0424 $\pm$ 0.000595   & 0.0219 $\pm$ 0.000712   & 0.2167 $\pm$ 0.002005   & 0.0345 $\pm$ 0.001032 \\
\midrule
\textbf{Helmholtz}   & 0.0519 $\pm$ 0.001751   & 0.0315 $\pm$ 0.002282   & 0.2310 $\pm$ 0.003499   & 0.0501 $\pm$ 0.002147 \\
\midrule
\begin{tabular}[l]{@{}l@{}}\textbf{Nonbounded}\\ \textbf{Navier-Stokes}\end{tabular} & 0.0368 $\pm$ 0.000786   & 0.0086 $\pm$ 0.000206   & 0.1097 $\pm$ 0.000403   & 0.0167 $\pm$ 0.000080 \\
\midrule
\textbf{Gray-Scott}  & 0.0393 $\pm$ 0.001163   & 0.0193 $\pm$ 0.000897   & 0.0382 $\pm$ 0.001187   & 0.0167 $\pm$ 0.000884 \\
\bottomrule
\end{tabular}%
}
\end{table}

Overall, the results show that SBNO is insensitive to the choice of inference seed. 
Across all datasets and both forward and inverse problems, the seed-level variation of the mean relative $L_2$ error is small compared with the corresponding test-set error level. These results indicate that the reported performance is not driven by a particular random seed. They also suggest that the across-instance error dispersion reported in the main results reflects variation in test-case difficulty rather than instability due to inference randomness.

\subsection{High-frequency Spectral Evaluation} \label{app:Spectral_Evaluation}

To examine whether reconstruction accuracy is accompanied by suppression of fine-scale structure, we evaluate high-frequency spectral error on Darcy flow and Helmholtz using the same 5\% observation protocol as in the main experiments. For each denormalized prediction $\hat y$ and ground truth $y$, we compute the orthonormal two-dimensional discrete Fourier transforms
$$ \widehat Y=\mathcal F(\hat y), \qquad Y=\mathcal F(y). $$
Using radial discrete frequency, we define
$$ \mathcal H = \left\{ \mathbf k: \|\mathbf k\|_2 \geq 0.5\max_{\mathbf j}\|\mathbf j\|_2 \right\}. $$
We stabilize normalization using
$$ D_{\mathrm{HF}}(y) = \max\left( \sum_{\mathbf k\in\mathcal H}|Y(\mathbf k)|^2,\; 10^{-12}\sum_{\mathbf k}|Y(\mathbf k)|^2+10^{-24} \right). $$
The high-frequency relative spectral error and energy ratio are
$$ \mathcal E_{\mathrm{HF}}(\hat y,y) = \sqrt{ \frac{ \sum_{\mathbf k\in\mathcal H} |\widehat Y(\mathbf k)-Y(\mathbf k)|^2 }{ D_{\mathrm{HF}}(y) } }, \quad \mathcal R_{\mathrm{HF}}(\hat y,y) = \frac{ \sum_{\mathbf k\in\mathcal H} |\widehat Y(\mathbf k)|^2 }{ D_{\mathrm{HF}}(y) }. $$
We compute both metrics per test instance and report their mean and standard deviation. All methods use the same Fourier normalization and frequency mask. Lower \(\mathcal E_{\mathrm{HF}}\) indicates more accurate recovery of the complex Fourier coefficients. When the denominator floor is inactive, \(\mathcal R_{\mathrm{HF}}=1\) indicates matching high-frequency energy, while values below or above one indicate attenuation or excess energy, respectively. Since the energy ratio does not measure phase or coefficient alignment, it must be interpreted jointly with the spectral error.

\begin{table}[htbp] 
\captionsetup{skip=1.5pt}
\caption{High-frequency relative spectral error under 5\% observations. Values are mean $\pm$ standard deviation across test instances; lower is better.} \label{table:hf-error}
\small
\centering
\resizebox{\linewidth}{!}{%
\begin{tabular}{lcccccccc}
\toprule
Method  & Darcy forward & Darcy inverse & Helmholtz forward & Helmholtz inverse \\
\midrule
\textbf{DiffusionPDE}  & 2.1465 $\pm$ 2.4595 & 1.2648 $\pm$ 0.1078 & 8.9174 $\pm$ 11.6318 & 0.4139 $\pm$ 0.2540 \\
\midrule
\textbf{FunDPS}     & 11.1638 $\pm$ 1.7811 & 1.3643 $\pm$ 0.0712 & 4.2907 $\pm$ 1.7299 & 0.5651 $\pm$ 0.1460 \\
\midrule
\textbf{SBNO}   & \textbf{1.6347 $\pm$ 0.1468} & \textbf{1.1543 $\pm$ 0.0621} & \textbf{1.9629 $\pm$ 1.1171} & \textbf{0.3278 $\pm$ 0.1533} \\
\bottomrule
\end{tabular}%
}
\end{table}

\begin{table}[htbp] 
\captionsetup{skip=1.5pt}
\caption{High-frequency energy ratios under 5\% observations. Values are mean $\pm$ standard deviation across test instances. A ratio of one indicates matching energy when the normalization floor is inactive; the ratios should be interpreted together with spectral error.} \label{table:hf-error}
\small
\centering
\resizebox{\linewidth}{!}{%
\begin{tabular}{lcccccccc}
\toprule
Method  & Darcy forward & Darcy inverse & Helmholtz forward & Helmholtz inverse \\
\midrule
\textbf{DiffusionPDE}  & 10.6676 $\pm$ 25.2757 & 0.9310 $\pm$ 0.3236 & 213.7049 $\pm$ 530.5397 & 0.9117 $\pm$ 0.2235 \\
\midrule
\textbf{FunDPS}     & 127.4999 $\pm$ 41.0276 & 1.1889 $\pm$ 0.1975 & 22.2814 $\pm$ 22.8030 & 0.9577 $\pm$ 0.2244 \\
\midrule
\textbf{SBNO}   & 2.6015 $\pm$ 0.5581 & 0.7429 $\pm$ 0.1215 & 6.2776 $\pm$ 8.0286 & 0.9932 $\pm$ 0.5402 \\
\bottomrule
\end{tabular}%
}
\end{table}

SBNO achieves the lowest high-frequency relative spectral error among the three evaluated methods in all four settings, reducing it relative to the next-best baseline by approximately 23.8\%, 8.7\%, 54.3\%, and 20.8\% for Darcy forward, Darcy inverse, Helmholtz forward, and Helmholtz inverse, respectively. The energy ratios do not support uniform high-frequency suppression as an explanation for these results: SBNO exhibits excess high-frequency energy in both forward tasks, although substantially less than the diffusion baselines. Some ground truths have little energy in this band, making relative energy ratios sensitive to small spurious components. Helmholtz inverse has a mean energy ratio close to one, whereas Darcy inverse exhibits attenuation. We therefore do not claim uniform spectral-energy preservation. Together, the two metrics indicate that SBNO's lower reconstruction-error dispersion cannot be attributed solely to high-frequency suppression in these evaluated settings.

\subsection{Mask-channel Ablation} \label{app:mask-ablation}

We assess the contribution of explicit observation-support information by independently training variants without the binary mask channel. These variants remove the mask coordinates from the lifted state and the inference ODE while retaining observed values, zero filling, randomly sampled sensor locations, the same backbone, and the 5\% observation protocol. Consequently, observed zeros and zero-filled missing entries are not explicitly distinguished in the no-mask representation.

\begin{table}[htbp] 
\captionsetup{skip=1.5pt}
\caption{Mask-channel ablation on all five benchmark datasets under 5\% observations. Values are mean $\pm$ standard deviation of relative $L_2$ error across test instances. For Gray–Scott, entries average the corresponding statistics of the two reconstructed channels.} \label{table:mask-ablation}
\small
\centering
\resizebox{\linewidth}{!}{%
\begin{tabular}{lcccccccc}
\toprule
Problem  & Mask-aware forward & No-mask forward & Mask-aware inverse & No-mask inverse \\
\midrule
\textbf{Darcy Flow}  & 0.0359 $\pm$ 0.0133 & 0.0579 $\pm$ 0.0172 & 0.0269 $\pm$ 0.0125 & 0.0297 $\pm$ 0.0138 \\
\midrule
\textbf{Poisson}     & 0.0424 $\pm$ 0.0219 & 0.3339 $\pm$ 0.1829 & 0.2167 $\pm$ 0.0345 & 0.3950 $\pm$ 0.1254 \\
\midrule
\textbf{Helmholtz}   & 0.0519 $\pm$ 0.0315 & 0.2411 $\pm$ 0.1428 & 0.2310 $\pm$ 0.0501 & 0.3334 $\pm$ 0.1149 \\
\midrule
\begin{tabular}[l]{@{}l@{}}\textbf{Nonbounded}\\ \textbf{Navier-Stokes}\end{tabular} & 0.0368 $\pm$ 0.0086 & 0.1471 $\pm$ 0.0238 & 0.1097 $\pm$ 0.0167 & 0.2267 $\pm$ 0.0243 \\
\midrule
\textbf{Gray-Scott}  & 0.0393 $\pm$ 0.0193 & 0.0448 $\pm$ 0.0248 & 0.0382 $\pm$ 0.0167 & 0.0385 $\pm$ 0.0210 \\
\bottomrule
\end{tabular}%
}
\end{table}

The mask-aware model has lower mean reconstruction error in all ten settings. On Darcy flow, removing the mask increases mean error by 61.3\% for forward reconstruction and 10.4\% for inverse reconstruction. Larger differences are observed on Poisson, Helmholtz, and nonbounded Navier–Stokes, whereas the difference for Gray–Scott inverse is small. These results support the utility of explicitly representing observation support, beyond supplying zero-filled values alone. The mask distinguishes genuine observed zeros from missing entries, but does not eliminate the non-uniqueness of reconstruction from partial observations.

\subsection{PDE-task Coupling Ablation} \label{app:pde-task_coupling_ablation}

We distinguish the direction of the PDE reconstruction task from the temporal direction of the bridge dynamics. The forward-task coupling $\Pi^F_{0,T}$ is induced by $(\mathcal E_s(a),\mathcal E_1(u))$, whereas the inverse-task coupling $\Pi^I_{0,T}$ is induced by $(\mathcal E_1(a),\mathcal E_s(u))$. We train Darcy variants using only $\Pi^F_{0,T}$, only $\Pi^I_{0,T}$, or their equal mixture. The dataset, backbone, 5\% observation setting, and evaluation protocol are otherwise unchanged. Both temporal drift networks are retained in every variant because the probability-flow ODE requires both drifts. Thus, task-specific training excludes one PDE-task coupling rather than one temporal network.

\begin{table}[htbp] 
\captionsetup{skip=1.5pt}
\caption{Mask-channel ablation on all five benchmark datasets under 5\% observations. Values are mean $\pm$ standard deviation of relative $L_2$ error across test instances. For Gray–Scott, entries average the corresponding statistics of the two reconstructed channels.} \label{table:mask-ablation}
\small
\centering
\resizebox{0.7\linewidth}{!}{%
\begin{tabular}{lcccccccc}
\toprule
Training coupling  & Forward error & Inverse error  \\
\midrule  
\textbf{Unified mixture}  & 0.0359 $\pm$ 0.0133 & 0.0269 $\pm$ 0.0125  \\
\midrule
\textbf{Forward-task coupling only}     & \textbf{0.0315 $\pm$ 0.0073} & 0.5063 $\pm$ 0.0525 \\
\midrule
\textbf{Inverse-task coupling only}   & 0.5420 $\pm$ 0.0269 & \textbf{0.0150 $\pm$ 0.0072} \\
\bottomrule
\end{tabular}%
}
\end{table}

Task-specific training reduces mean error on the retained task by approximately 12.3\% for forward reconstruction and 44.2\% for inverse reconstruction relative to unified training. However, error on the excluded task increases substantially: forward-task-only training yields an inverse error of 0.5063, while inverse-task-only training yields a forward error of 0.5420. These results reveal a specialization–unification trade-off. In this Darcy experiment, adding the opposite PDE-task coupling does not improve the retained task's point-reconstruction accuracy. The benefit of unified training is that one checkpoint supports both PDE tasks, whereas specialization improves the selected task at the expense of the unsupported capability.

\subsection{Forward-drift-only SDE Evaluation} \label{app:forward-drift-only_SDE_evaluation}

The two temporal drift networks have separate parameters, losses, optimizers, and exponential moving averages. Neither network uses the other network's outputs or parameters during optimization. Therefore, the forward drift extracted from a unified checkpoint has the same training objective as a forward drift trained alone on the same coupling mixture, although separately executed training runs need not yield identical parameters.

To evaluate the forward drift without the backward drift, we use forward SDE inference, since our probability-flow ODE requires both networks. We compare the forward drift trained on the unified mixture with that trained only on the forward-task coupling. Each test case uses 60 stochastic paths, with identical observation masks and Brownian seeds across the two variants. We evaluate the sample-mean reconstruction and report the mean and standard deviation of its relative $L_2$ error across test instances.

\begin{table}[htbp] 
\captionsetup{skip=1.5pt}
\caption{Mask-channel ablation on all five benchmark datasets under 5\% observations. Values are mean $\pm$ standard deviation of relative $L_2$ error across test instances. For Gray–Scott, entries average the corresponding statistics of the two reconstructed channels.} \label{table:mask-ablation}
\small
\centering
\resizebox{0.7\linewidth}{!}{%
\begin{tabular}{lcccccccc}
\toprule
Forward-drift training coupling  & Inference & Forward error  \\
\midrule  
\textbf{Unified mixture}  & Forward-drift-only SDE & 0.0263 $\pm$ 0.0064  \\
\midrule
\textbf{Forward-task coupling only}     & Forward-drift-only SDE & \textbf{0.0174 $\pm$ 0.0056} \\
\bottomrule
\end{tabular}%
}
\end{table}

Specializing the forward drift to the forward-task coupling reduces the mean error by approximately 34\%. Consistent with the ODE coupling ablation, this matched SDE comparison does not show an accuracy benefit from adding the opposite-task coupling on Darcy forward reconstruction. Unified training instead provides both PDE capabilities when the two learned drifts are used together. Because this table evaluates a 60-path sample mean, its error values should not be interpreted as a comparison of individual SDE samples with deterministic ODE predictions.

\section{Additional Experiments on Diverse Datasets}
\label{app:additional_experiments}

In this section, we report additional experimental results on three time-dependent datasets that were not included in the main text: Korteweg--De Vries, Kuramoto--Sivashinsky, and Kolmogorov flow. Details of these datasets are provided in Appendix~\ref{app:datasets}. Unless otherwise stated, all experimental settings, training protocols, and evaluation procedures are the same as those used in Section~\ref{sec:experiments}.

We separate these experiments from the main text because their scale is substantially smaller than that of the primary benchmarks in Section~\ref{sec:experiments}. In particular, these datasets contain fewer trajectories and fewer total training pairs, making them less suitable as the main basis for large-scale quantitative comparison. Nevertheless, they provide a useful complementary testbed, since they exhibit different forms of nonlinear and turbulent dynamics from those considered in the main experiments. In this sense, they offer additional evidence on whether SBNO can remain effective beyond the specific PDE families emphasized in Section~\ref{sec:experiments}.

\begin{table*}[ht]
\captionsetup{skip=1.5pt}
\caption{Evaluation results on three additional APEBench datasets. Each performance metric is reported as the ``mean $\pm$ standard deviation'' of the relative $L_2$ error across test samples. For clarity, the best mean and best standard deviation in each row are highlighted in \textbf{bold}, while the second-best values are marked with \underline{underline}.}
\label{table:Additional_experiments_result}
\small
\centering
\resizebox{1.0\textwidth}{!}{%
\begin{tabular}{llcccccc}
\toprule
\textbf{Dataset} & \textbf{Problem} & \textbf{SBNO (Ours)} & \textbf{FunDPS} & \textbf{DiffusionPDE} & \textbf{FNOPE} & \textbf{GNOT} & \textbf{FNO} \\
\midrule
\multirow{2}{*}{\shortstack[l]{Korteweg\\–De Vries}} 
& Forward & \underline{0.2858} $\pm$ \underline{0.0650} &  1.5024 $\pm$ 0.2236 &  \textbf{0.2177} $\pm$ 0.2017 &  0.8498 $\pm$ 0.0446  &  1.1628 $\pm$ 0.0692              &  0.5356 $\pm$ \textbf{0.0348}  \\
& Inverse & \underline{0.2866} $\pm$ 0.1005             &  1.5401 $\pm$ 0.2491 &  \textbf{0.1765} $\pm$ 0.0791 &  0.9399 $\pm$ \underline{0.0571}  &  1.1798 $\pm$ 0.0739  &  0.7483 $\pm$ \textbf{0.0202} \\
\midrule
\multirow{2}{*}{\shortstack[l]{Kuramoto–\\Sivashinsky}} 
& Forward & \textbf{0.3236} $\pm$ \textbf{0.0295} &  1.2990 $\pm$ 0.0754 &  \underline{0.4517} $\pm$ 0.0992 &  1.0500 $\pm$ 0.0711  &  1.0707 $\pm$ 0.0579 &  0.6808 $\pm$ \underline{0.0359}  \\
& Inverse & \textbf{0.3674} $\pm$ \textbf{0.0283} &  1.2962 $\pm$ 0.0733 &  \underline{0.3973} $\pm$ 0.0584 &  1.0451 $\pm$ 0.0716  &  1.0660 $\pm$ 0.0568 &  0.6727 $\pm$ \underline{0.0389} \\
\midrule
\multirow{2}{*}{\shortstack[l]{Kolmogorov \\Flow}} 
& Forward & \textbf{0.5322} $\pm$ 0.0978 &  1.2445 $\pm$ \textbf{0.0560} &  \underline{0.5551} $\pm$ 0.1133 &  0.6180 $\pm$ 0.0882  &  0.9275 $\pm$ 0.0860 &  0.6211 $\pm$ \underline{0.0764}  \\
& Inverse & \textbf{0.4849} $\pm$ 0.0810 &  1.2329 $\pm$ \underline{0.0525} &  \underline{0.5739} $\pm$ 0.2634 &  0.5687 $\pm$ 0.0632  &  0.9071 $\pm$ 0.0814 &  0.6039 $\pm$ \textbf{0.0466} \\
\bottomrule
\end{tabular}%
}
\end{table*}

Table~\ref{table:Additional_experiments_result} summarizes the reconstruction results under the same partial-observation setting as in the main paper. Overall, SBNO remains competitive across all three datasets and achieves the best mean error on four out of the six forward/inverse tasks, namely both Kuramoto-Sivashinsky tasks and both Kolmogorov flow tasks. On Korteweg-de Vries, SBNO achieves the second-best mean error in both directions, while DiffusionPDE performs best. This trend is consistent with our observations in some of the more ill-posed settings in the main text: PDE-guided diffusion baselines can benefit from additional physics-based constraints at inference time, whereas SBNO is intentionally equation-free. Even in these cases, however, SBNO remains substantially more accurate than the other deterministic and probabilistic operator baselines considered here.

Another notable trend is that SBNO often combines competitive mean accuracy with relatively low dispersion, although the smallest standard deviation is not always achieved by SBNO. In several cases, FNO attains a smaller error standard deviation despite having a clearly worse mean error, indicating more concentrated but systematically less accurate predictions. By contrast, the diffusion baselines sometimes exhibit strong average performance but also larger variability across test samples. Taken together, these results suggest that SBNO continues to provide a favorable trade-off between reconstruction accuracy and test-instance error dispersion even on smaller-scale datasets with distinct dynamical characteristics.


\section{Uncertainty Quantification with Stochastic Sampling} \label{app:uncertainty_quantification}

Although the main paper focuses on deterministic inference via the probability-flow ODE for efficiency and stable deterministic point reconstruction, the learned fixed-coupling bridge dynamics in SBNO can also be used in a stochastic sampling mode for uncertainty quantification. To examine this capability more broadly, we additionally perform stochastic-sampling experiments for all datasets and for both forward and inverse problems under the same partial-observation setting as in the main experiments. Specifically, for each test case, we draw 60 stochastic samples from the learned bridge dynamics and report their empirical predictive statistics in Table~\ref{table:uncertainty_quantification}, without assuming exact recovery of the target conditional law.

\begin{table}[htbp] 
\captionsetup{skip=1.5pt}
\caption{Uncertainty-quantification results of SBNO using stochastic posterior sampling. For each dataset and for both forward and inverse problems, we draw 60 stochastic bridge samples per test case and report metrics for sample-mean reconstruction, uncertainty magnitude, calibration and sharpness, and the relation between predictive error and estimated uncertainty. \label{table:uncertainty_quantification}}
\small
\centering
\resizebox{\linewidth}{!}{%
\begin{tabular}{llcccccccccc}
\toprule
\multirow{2}{*}{Category}  & \multirow{2}{*}{Metric}  & \multicolumn{2}{c}{\textbf{Darcy Flow}} & \multicolumn{2}{c}{\textbf{Poisson}} & \multicolumn{2}{c}{\textbf{Helmholtz}} & \multicolumn{2}{c}{\begin{tabular}[c]{@{}c@{}}\textbf{Nonbounded}\\ \textbf{Navier-Stokes}\end{tabular}} & \multicolumn{2}{c}{\textbf{Gray-Scott}} \\ 
\cmidrule(lr){3-4} \cmidrule(lr){5-6} \cmidrule(lr){7-8} \cmidrule(lr){9-10} \cmidrule(lr){11-12}
& & Forward & Inverse & Forward & Inverse & Forward & Inverse & Forward & Inverse & Forward & Inverse \\
\midrule
\multirow{3}{*}{\begin{tabular}[l]{@{}l@{}}Sample mean\\ reconstruction\end{tabular}} & Rel $L_2$ error mean  & 0.0263  & 0.1296  & 0.0519  & 0.1271  & 0.0539  & 0.1362  & 0.0717  & 0.1342  & 0.0254 & 0.0272  \\
          & Rel $L_2$ error std   & 0.0064  & 0.0334  & 0.0210  & 0.0361  & 0.0297  & 0.0362  & 0.0072  & 0.0182  & 0.0119  & 0.0162  \\
          & Abs $L_1$ error mean  & 0.0001  & 0.2217  & 0.0001  & 0.0257  & 0.0002  & 0.0260  & 0.0142  & 0.0267  & 0.0089  & 0.0037  \\
\midrule
\multirow{2}{*}{\begin{tabular}[l]{@{}l@{}}Uncertainty \\ magnitude\end{tabular}} &  Global mean std  & 0.0002  & 0.2016  & 0.0001  & 0.0144  & 0.0001  & 0.0131  & 0.0128  & 0.0172  & 0.0116  & 0.0060  \\
   &  Global mean var  & 0.0000  & 0.6258  & 0.0000  & 0.0002  & 0.0000  & 0.0002  & 0.0002  & 0.0003  & 0.0002  & 0.0001  \\
\midrule
\multirow{6}{*}{\begin{tabular}[l]{@{}l@{}}Calibration \\ and sharpness\end{tabular}}  & 50\% Coverage & 0.2826  & 0.9858  & 0.2601  & 0.2170  & 0.2438  & 0.2025  & 0.1895  & 0.1602  & 0.3946  & 0.5085  \\
   & 80\% Coverage & 0.5425  & 0.9903  & 0.4949  & 0.3923  & 0.4632  & 0.3712  & 0.4078  & 0.3150  & 0.6617  & 0.8169  \\
   & 90\% Coverage & 0.6987  & 0.9923  & 0.6241  & 0.4806  & 0.5843  & 0.4574  & 0.6256  & 0.4679  & 0.7697  & 0.9136  \\
   & 50\% Width    & 0.0001  & 0.1524  & 0.0001  & 0.0181  & 0.0001  & 0.0167  & 0.0100  & 0.0141  & 0.0117  & 0.0062  \\
   & 80\% Width    & 0.0003  & 0.2985  & 0.0002  & 0.0350  & 0.0002  & 0.0321  & 0.0212  & 0.0292  & 0.0246  & 0.0128  \\
   & 90\% Width    & 0.0005  & 0.4012  & 0.0004  & 0.0451  & 0.0003  & 0.0413  & 0.0362  & 0.0491  & 0.0336  & 0.0174  \\
\midrule
\begin{tabular}[l]{@{}l@{}}Error-uncertainty \\ relation\end{tabular}  &  Corr(error, std)  & 0.2362  & 0.6181  & 0.4778  & 0.2103  & 0.0908  & 0.2315  & 0.0362  & 0.0814  & 0.4720  & 0.5787  \\
\bottomrule
\end{tabular}%
}
\end{table}

We organize the reported quantities into four groups. First, under ``sample-mean reconstruction'', we evaluate the accuracy of the sample mean prediction using the relative $L_2$ error mean and standard deviation across test samples, together with the mean absolute $L_1$ error. These metrics indicate whether stochastic sampling preserves accurate reconstruction when the posterior is summarized by its mean. Second, under ``uncertainty magnitude'', we report the global mean sample standard deviation and variance, which quantify the overall scale of predictive uncertainty across the spatial domain and test set.

Third, under ``calibration and sharpness'', we report empirical coverage and interval width at the 50\%, 80\%, and 90\% empirical predictive intervals. The empirical coverage measures how often the ground-truth field values fall inside the corresponding posterior interval, while the interval width reflects the sharpness of the uncertainty estimate. Taken together, these metrics characterize whether the empirical predictive distribution is both informative and appropriately calibrated: wider intervals may improve coverage but reduce sharpness, whereas overly narrow intervals may indicate overconfident uncertainty estimates.

Finally, under ``error--uncertainty relation'', we report the correlation between the absolute error of the sample mean and the sample standard deviation. This correlation measures the spatial association between reconstruction error and sample variability; its strength varies substantially across datasets and tasks.

\begin{figure} [ht]
    \centering
    \includegraphics[width=0.9\columnwidth]{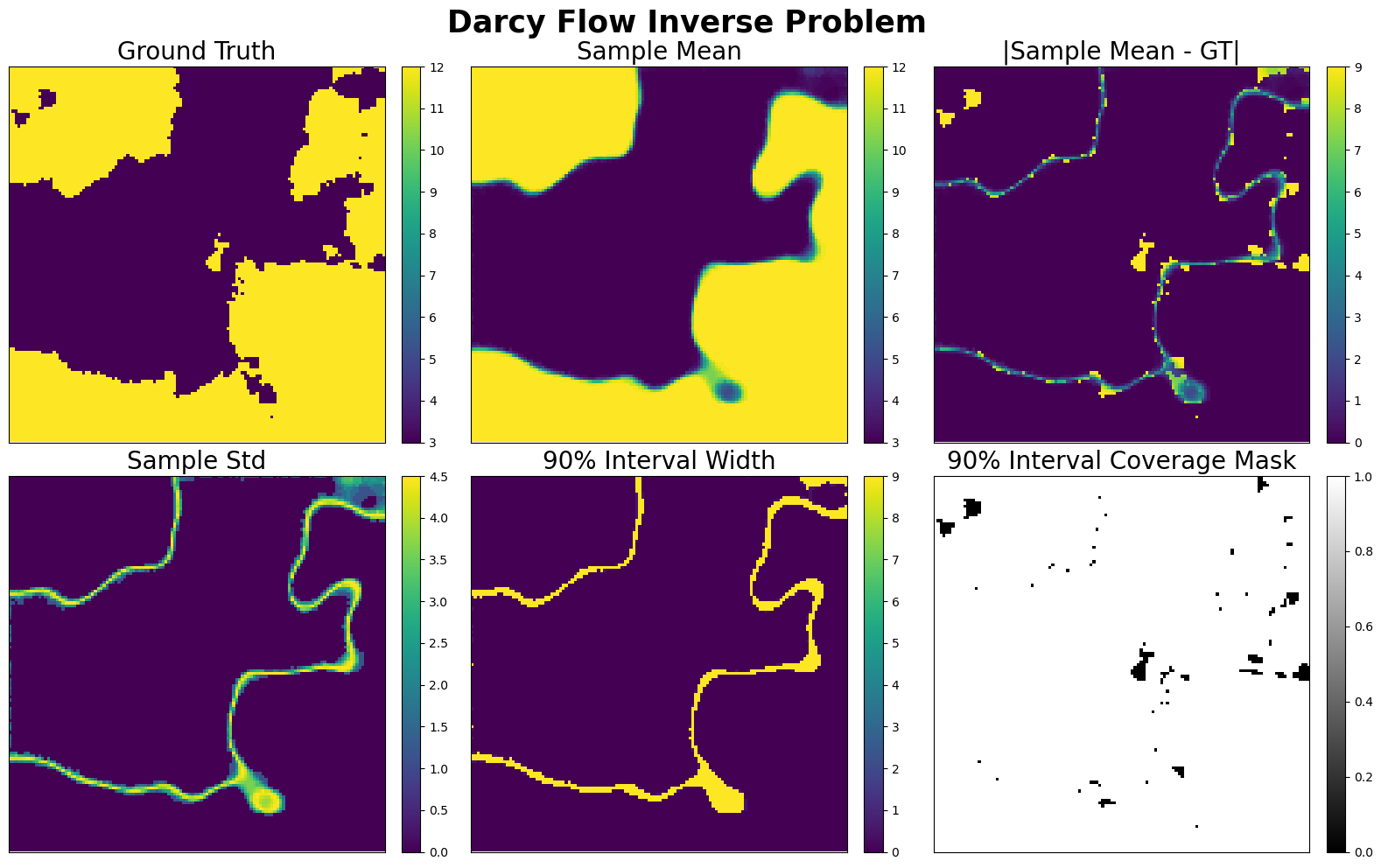}
    \caption{Representative uncertainty-quantification example of SBNO under partial observations using stochastic posterior sampling. We draw 60 stochastic bridge samples and visualize the ground truth, sample mean, absolute error of the sample mean, sample standard deviation, 90\% posterior interval width, and the corresponding 90\% empirical coverage mask. The sample mean summarizes the reconstruction, while the posterior spread highlights spatial regions associated with larger predictive uncertainty.}
    \label{fig:uncertainty}
\end{figure}

Overall, these metrics provide a complementary view to the deterministic results in the main paper. Deterministic probability-flow ODE inference remains preferable when a fast and stable deterministic point reconstruction is desired, whereas stochastic bridge sampling enables empirical predictive-uncertainty estimates that can be evaluated in terms of reconstruction quality, uncertainty magnitude, calibration, and error awareness. For illustration, Figure~\ref{fig:uncertainty} visualizes a representative example of the empirical predictive distribution, including the sample mean, sample standard deviation, interval width, and empirical coverage mask.

\begin{figure} [ht]
    \centering
    \includegraphics[width=0.9\columnwidth]{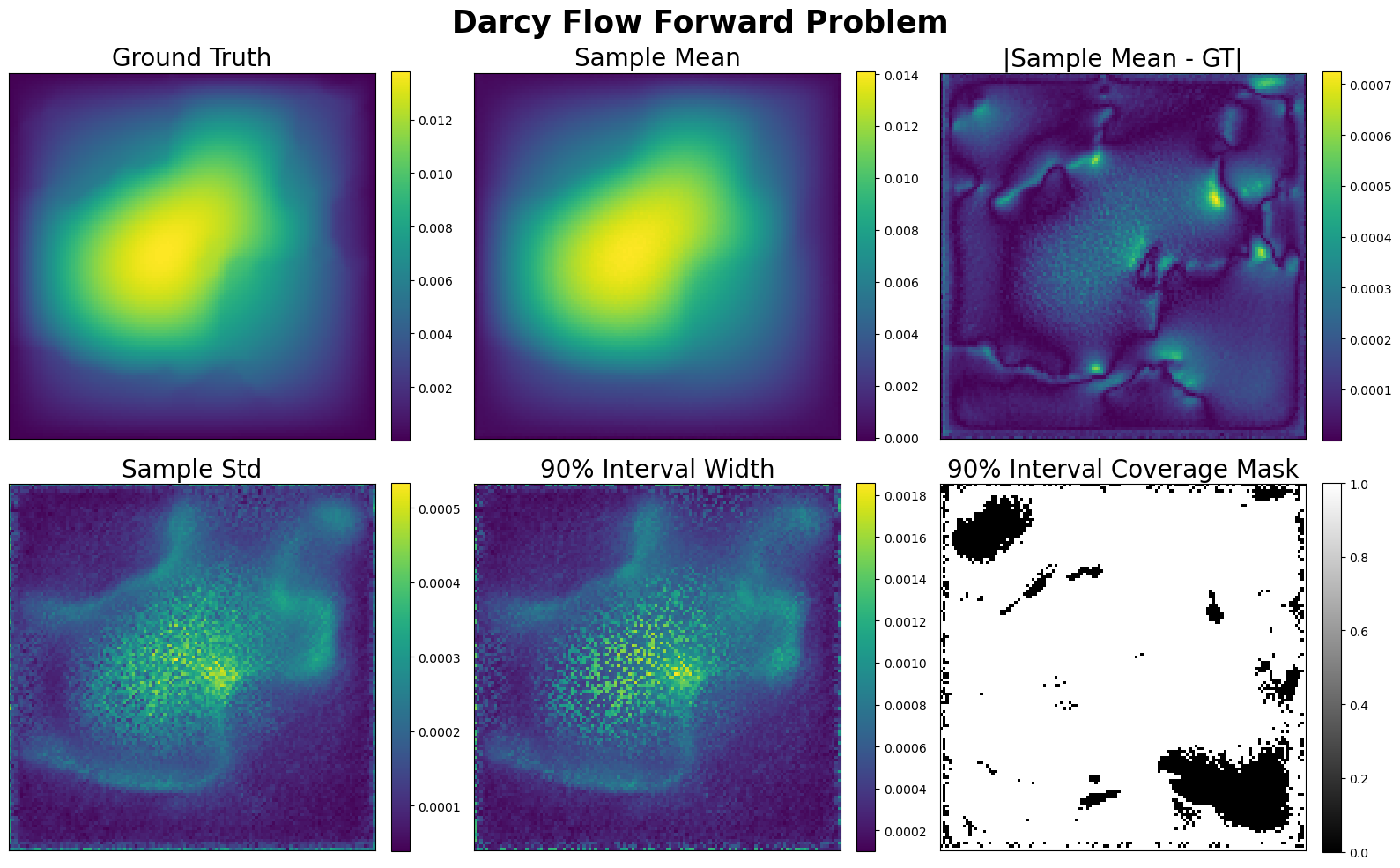}
\end{figure}

\begin{figure} [ht]
    \centering
    \includegraphics[width=0.8\columnwidth]{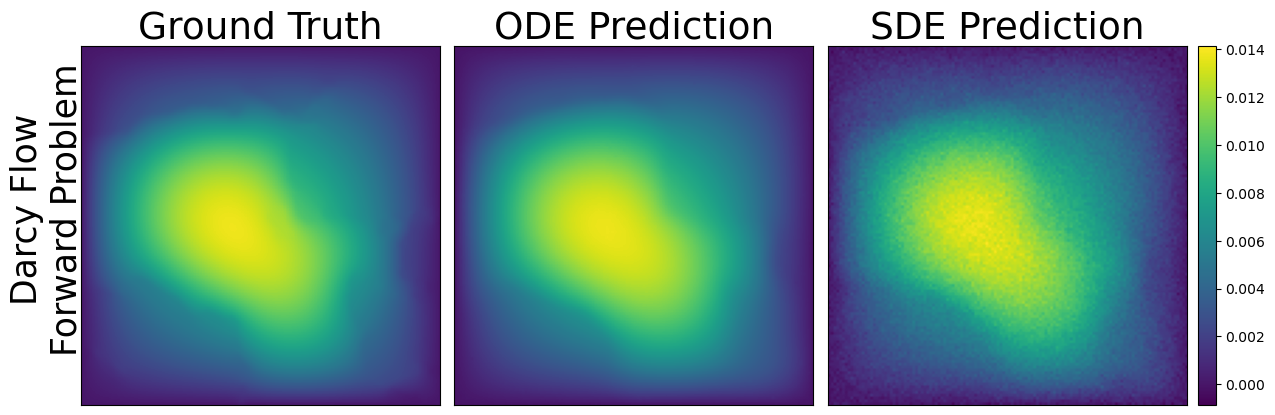}
    \includegraphics[width=0.8\columnwidth]{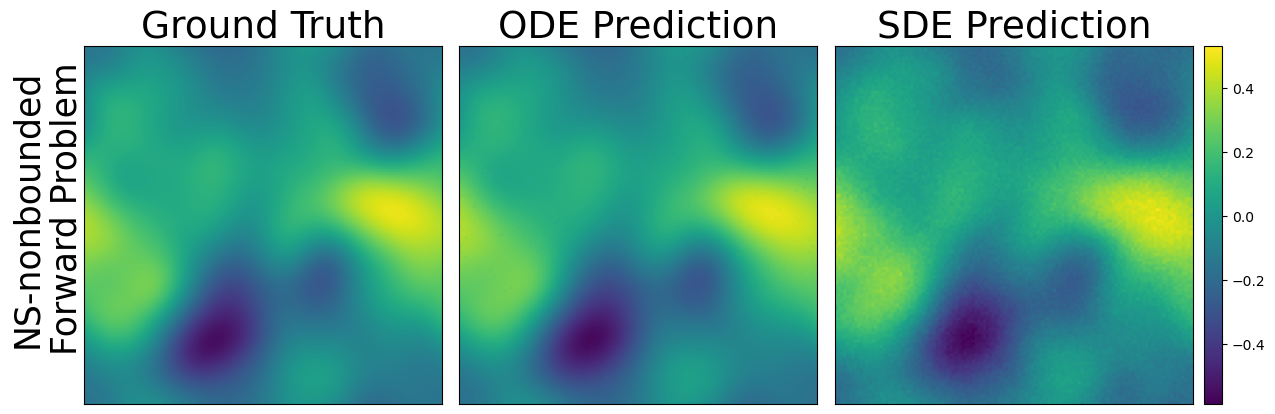}
    \caption{Comparison of ODE and SDE inference on representative Darcy-flow and nonbounded Navier–Stokes forward cases under 5\% observations. For each case, we show the ground truth, deterministic ODE reconstruction, one stochastic SDE sample, and the mean of 60 SDE samples. All predictions use the same observation mask, and panels within each case share a color scale. The individual SDE sample is shown separately from the sample-mean reconstruction.}
    \label{fig:ODE_SDE_Comparison}
\end{figure}

In figure~\ref{fig:ODE_SDE_Comparison}, we compare ODE reconstruction with an individual SDE sample on representative Darcy-flow and nonbounded Navier–Stokes forward cases. In these examples, ODE inference reproduces the dominant spatial structures, while individual SDE predictions retain the overall pattern but exhibit more diffuse features and reduced sharpness. The individual sample should be distinguished from the 60-sample mean evaluated in Table~\ref{table:uncertainty_quantification}. These visualizations illustrate the behavior of the two inference modes on selected cases; they do not establish calibrated uncertainty or exact recovery of the target conditional distribution.

\section{Robustness to Noisy Partial Observations} \label{app:noisy_observation}

We additionally evaluate the robustness of SBNO under noisy partial observations, which are common in practical measurement settings. To this end, we conduct additional experiments on Darcy flow by injecting Gaussian noise into the observed partial inputs at multiple noise levels. We compare SBNO against the diffusion-based baselines under the same protocol, and the corresponding results are summarized in Table~\ref{table:noisy_observation}.

\begin{table*}[ht]
\captionsetup{skip=1.5pt}
\caption{Evaluation results with Gaussian noise added to partial observations. Noise std denotes the standard deviation of the Gaussian noise. For clarity, the best mean in each column is highlighted in \textbf{bold}, while the second-best values are marked with \underline{underline}.}
\label{table:noisy_observation}
\small
\centering
\resizebox{0.7\textwidth}{!}{%
\begin{tabular}{llcccc}
\toprule
\textbf{Noise std} & \textbf{Problem} & \textbf{SBNO (Ours)} & \textbf{FunDPS} & \textbf{DiffusionPDE}  & \textbf{GNOT}  \\
\midrule
\multirow{2}{*}{0.1} & Forward & \textbf{0.0359}  &  0.0461  &  \underline{0.0371} & 0.0510 \\
                     & Inverse & \textbf{0.0269}  &  0.0834  &  \underline{0.0743} & 0.1031 \\
\midrule
\multirow{2}{*}{0.5} & Forward & \textbf{0.0359}  &  0.0461  &  \underline{0.0372} & 0.1709 \\
                     & Inverse & \textbf{0.0274}  &  0.0837  &  \underline{0.0749} & 0.4092 \\
\midrule
\multirow{2}{*}{1.0} & Forward & \textbf{0.0358}  &  0.0461  &  \underline{0.0373} & 0.2657 \\
                     & Inverse & \textbf{0.0291}  &  0.0838  &  \underline{0.0766} & 0.4764 \\
\midrule
\multirow{2}{*}{3.0} & Forward & \textbf{0.0363}  &  0.0461  &  \underline{0.0382} & 0.3389 \\
                     & Inverse & \textbf{0.0448}  &  0.0861  &  \underline{0.0851} & 0.5011 \\
\bottomrule
\end{tabular}%
}
\end{table*}

As shown in Table~\ref{table:noisy_observation}, SBNO remains competitive across all tested noise levels in both the forward and inverse problems. Under low to moderate noise, the reconstruction quality changes only slightly, indicating that the model is reasonably stable to imperfect observations in this regime. As the noise level increases further, performance degrades as expected, with the effect being more noticeable in the inverse problem than in the forward problem. This trend is consistent with the greater ill-posedness of inverse inference under sparse and corrupted observations. Nevertheless, SBNO continues to compare favorably with the considered baselines throughout the evaluated range.

These results strengthen the empirical evaluation by showing that SBNO is not limited to the clean-observation setting. At the same time, they also indicate that robustness under noisy measurements remains an important dimension of the problem, especially in more challenging inverse settings. Other forms of measurement corruption and non-Gaussian noise remain important directions for future work.


\section{Sensitivity to the ODE Solver and Number of Inference Steps} \label{app:ODE_solver_step}
We additionally investigate whether the performance gap on the more challenging inverse tasks can be reduced by using a more accurate ODE solver configuration at inference time. To this end, we evaluate SBNO with increased numbers of RK4 steps and also consider an adaptive DOPRI solver. We focus on the two inverse settings where SBNO did not outperform DiffusionPDE in the main experiments. The corresponding reconstruction performance and per-sample inference time are summarized in Table~\ref{table:ODE_solver}.

\begin{table*}[ht]
\captionsetup{skip=1.5pt}
\caption{Reconstruction performance under different ODE solver configurations on the challenging inverse tasks where SBNO did not outperform DiffusionPDE in the main text, together with the corresponding per-sample inference time. We increase the number of RK4 integration steps and additionally evaluate an adaptive DOPRI solver to assess whether improved numerical integration can close the remaining performance gap.}
\label{table:ODE_solver}
\small
\centering
\resizebox{0.75\textwidth}{!}{%
\begin{tabular}{lccccc}
\toprule
\textbf{Step number} & 32  & 64 & 96 & 128 & dopri \\
\midrule
Poisson Inverse    & 0.2167 & 0.2074 & 0.2072 & 0.2073 & 0.2166   \\
Helmholtz Inverse  & 0.2310 & 0.2220 & 0.2243 & 0.2249 & 0.2310    \\
\midrule
Time per Sample(sec) & 0.28 & 0.56 & 0.83 & 1.10 & 0.57    \\
\bottomrule
\end{tabular}%
}
\end{table*}

As shown in Table~\ref{table:ODE_solver}, increasing the number of ODE solver steps leads to a modest improvement over the default inference configuration, indicating that part of the error can be reduced through finer numerical integration. However, the gains quickly saturate as the step count increases, and the overall improvement remains limited relative to the additional inference cost. A similar trend is observed with the adaptive solver, which does not provide a meaningful advantage over the fixed-step configurations. Overall, these results suggest that the remaining performance gap on these tasks is not primarily caused by insufficient numerical accuracy of the probability-flow ODE solver.

This observation is consistent with our explanation for the challenging inverse settings. In these cases, the dominant difficulty appears to come from the ambiguity of the inverse mapping under sparse observations rather than from discretization error along the deterministic flow trajectory. As a result, simply refining the deterministic solver is not sufficient to close the gap. This analysis clarifies that the default inference setting provides a practical trade-off between reconstruction quality and efficiency, while more aggressive solver settings yield only limited additional benefit.


\section{Sensitivity to the Observation Ratio}
\label{app:observation_ratio}

We additionally evaluate how SBNO behaves when the amount of available partial observation at inference time differs from the 5\% regime used in the main experiments. To this end, we vary the observation ratio at test time while keeping the training setup unchanged, and report the corresponding reconstruction performance in Table~\ref{table:observation_ratio}.

\begin{table*}[ht]
\captionsetup{skip=1.5pt}
\caption{Evaluation results with varying ratios of partial observations. The observation ratio denotes the proportion of points in the full grid that are provided as partial observations for evaluation.}
\label{table:observation_ratio}
\small
\centering
\resizebox{0.75\textwidth}{!}{%
\begin{tabular}{lcccccc}
\toprule
\textbf{Observation ratio} & \textbf{5\%} & \textbf{4\%}  & \textbf{3\%} & \textbf{2\%} & \textbf{1\%} & \textbf{0.5\%} \\
\midrule
Forward & 0.0359 & 0.0402 & 0.0749 & 0.1226 & 0.2212 & 0.2727   \\
Inverse & 0.0269 & 0.0498 & 0.0865 & 0.2107 & 0.2924 & 0.3362   \\
\bottomrule
\end{tabular}%
}
\end{table*}

As shown in Table~\ref{table:observation_ratio}, SBNO performs best at the 5\% observation ratio used during training, and the reconstruction error increases as the number of observed points decreases. This trend is visible in both the forward and inverse problems, with the degradation becoming more pronounced in the lower-observation regime. In particular, once the observation ratio falls substantially below the training condition, performance drops more sharply, especially for the inverse problem.

This behavior is consistent with a shift in the conditioning distribution induced by changing the observation ratio. In the current setup, SBNO is optimized using masked observations sampled to match the target sparse setting in the main experiments, namely 5\% observations. When inference is instead performed with substantially fewer observations, the model must extrapolate to a much weaker conditioning signal than it was exposed to during training. As a result, the learned deterministic reconstruction becomes less reliable as the available information decreases.

These results clarify an important limitation of the current formulation. While SBNO is effective in the sparse-observation regime it is trained for, its performance is not yet robust across substantially different observation ratios without retraining or additional adaptation.


\section{Dynamic-ratio Training}
Dynamic-ratio training. To examine whether exposure to varying conditioning densities improves robustness, we train a separate Darcy model with $r\sim\mathrm{Uniform}(0.5\%,10\%)$ at each training iteration, while also resampling sensor locations. We then evaluate the same checkpoint across the trained ratio range without retraining or adaptation.

\begin{table*}[ht]
\captionsetup{skip=1.5pt}
\caption{Evaluation results with varying ratios of partial observations. The observation ratio denotes the proportion of points in the full grid that are provided as partial observations for evaluation.}
\label{table:observation_ratio}
\small
\centering
\resizebox{0.75\textwidth}{!}{%
\begin{tabular}{lcccccccc}
\toprule
\textbf{Observation ratio} & \textbf{10\%} & \textbf{8\%} & \textbf{5\%} & \textbf{4\%}  & \textbf{3\%} & \textbf{2\%} & \textbf{1\%} & \textbf{0.5\%} \\
\midrule
Forward & 0.0664 & 0.0703 & 0.0836 & 0.0880 & 0.0954 & 0.1112 & 0.1435 & 0.1790   \\
Inverse & 0.0442 & 0.0437 & 0.0474 & 0.0511 & 0.0595 & 0.0773 & 0.1240 & 0.1710   \\
\bottomrule
\end{tabular}%
}
\end{table*}

Forward error decreases throughout the evaluated range, while inverse error decreases and approximately plateaus at 8\%–10\%. Compared with the fixed-density model, dynamic training reduces forward/inverse errors from 0.2727/0.3362 to 0.1790/0.1710 at 0.5\%, and from 0.1266/0.0810 to 0.0664/0.0442 at 10\%. However, at the specialized 5\% condition, the fixed-density model remains more accurate, with errors of 0.0359/0.0269 compared with 0.0836/0.0474. These results demonstrate a specialization–robustness trade-off: fixed-density training favors a known observation regime, whereas dynamic-ratio training accommodates varying sensor availability within the evaluated range. We have not evaluated the dynamic model at 15\% or 20\%, so these results do not establish robustness beyond its training range.


\section{Closed-loop Forward/Inverse Evaluation}
To examine recursive use, we perform an evaluation-only closed-loop experiment on held-out pairs with independently sampled observation masks, without closed-loop training. Specifically, after obtaining a prediction in one direction, we reuse that prediction as the partial observation for inference in the opposite direction. Concretely, for forward inference, we use the prediction from backward inference as the partial input observation, and for backward inference, we use the prediction from forward inference as the partial solution observation. Training uses clean partial fields rather than model-generated feedback, and includes neither closed-loop supervision nor a cycle-consistency objective. We did not evaluate training-set closed-loop performance, so the reported results do not establish greater consistency on training pairs. We evaluate this closed-loop setting on Darcy flow and Helmholtz. We choose Darcy flow because SBNO performs strongly in both directions on this dataset and Helmholtz because it provides a more challenging case, particularly for the inverse problem. The evaluation results are presented in Table~\ref{table:loop_results}. We denote the resulting reconstructions as SBNO (loop).

\begin{table*}[ht]
\captionsetup{skip=1.5pt}
\caption{Evaluation results for the closed-loop experiment, where inference is run again in the opposite direction using the model prediction from the first pass as the new partial observation. For forward inference, the prediction obtained from backward inference is used as the partial input observation. Conversely, for backward inference, the prediction obtained from forward inference is used as the partial solution observation. The resulting performance is reported as ``SBNO (loop).''}
\label{table:loop_results}
\small
\centering
\resizebox{0.7\textwidth}{!}{%
\begin{tabular}{llcc}
\toprule
\textbf{Dataset} & \textbf{Problem} & \textbf{SBNO} & \textbf{SBNO (loop)} \\
\midrule
\multirow{2}{*}{Darcy flow} & Forward & 0.0359 $\pm$ 0.0133 &  0.0369 $\pm$ 0.0132  \\
                            & Inverse & 0.0269 $\pm$ 0.0128 &  0.0531 $\pm$ 0.0216   \\
\midrule
\multirow{2}{*}{Helmholtz}  & Forward & 0.0519 $\pm$ 0.0315 &  0.1886 $\pm$ 0.1069  \\
                            & Inverse & 0.2310 $\pm$ 0.0501 &  0.2506 $\pm$ 0.0722  \\
\bottomrule
\end{tabular}%
}
\end{table*}

On Darcy flow, the forward problem remains largely stable under looping, with the relative $L_2$ error changing only slightly. In contrast, the inverse problem shows a more noticeable degradation. On Helmholtz, the degradation is more pronounced in the forward direction, while the inverse direction changes more moderately.

\begin{figure}[ht]
    \centering
    \includegraphics[width=\linewidth]{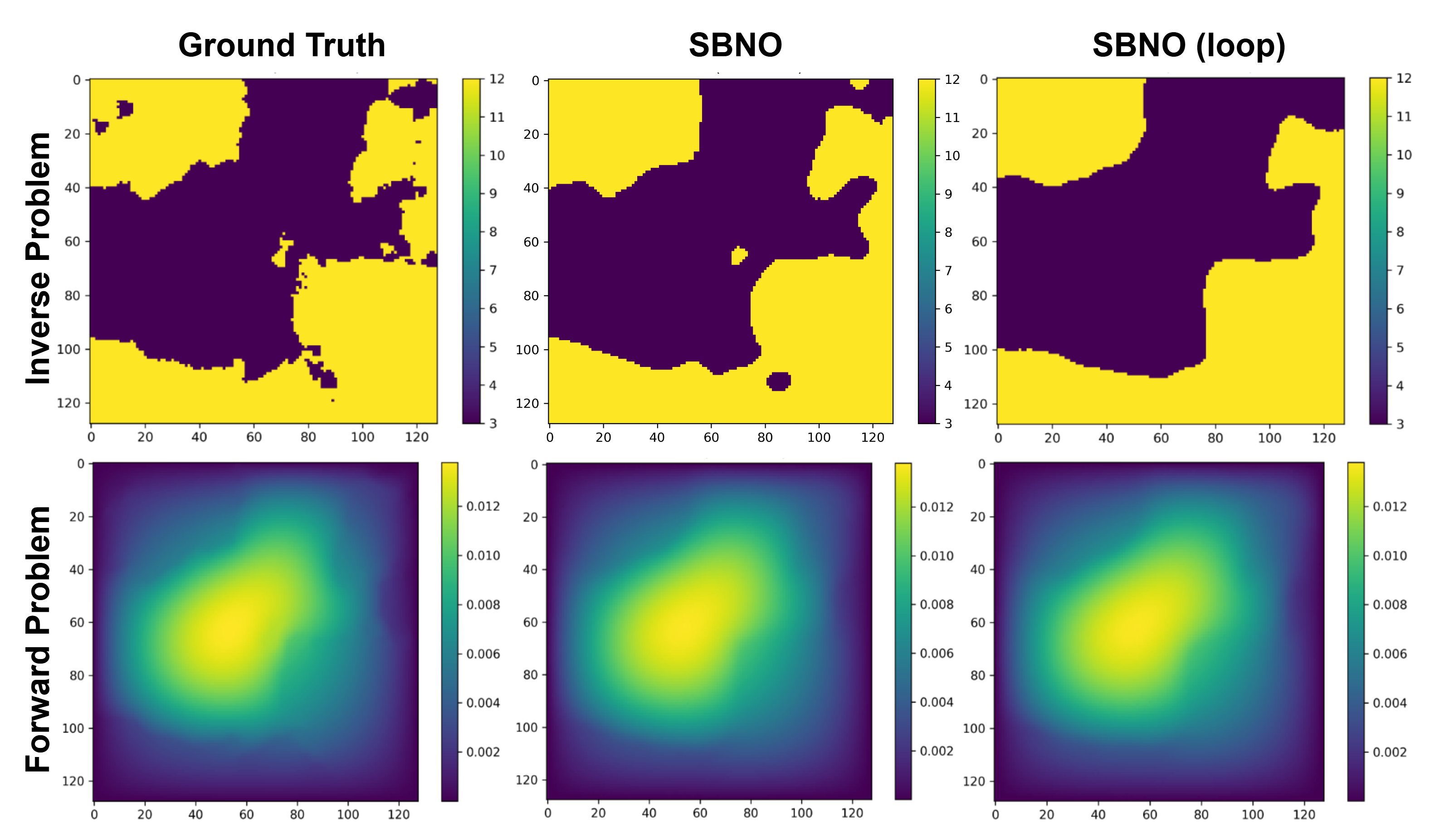}
    \caption{\textbf{Representative closed-loop visualization}. We show the original SBNO predictions and the reconstructions obtained after reusing those predictions as partial observations for inference in the opposite direction.}
    \label{fig:looping}
\end{figure}

Overall, the closed-loop experiment indicates that SBNO preserves a meaningful degree of directional compatibility, but does not fully eliminate recursive error accumulation. In particular, once model predictions are reused as observations, they no longer behave like clean partial measurements and instead resemble noisy or smoothed surrogates of the true field, already containing approximation error and attenuated high-frequency structure. As a result, repeated forward/inverse reuse tends to produce smoother reconstructions with reduced fine details, together with increased reconstruction error in the harder settings. This can be seen in the Figure~\ref{fig:looping}. These observations indicate limited directional compatibility in the unified model, with recursive reconstruction remaining sensitive to errors in the intermediate prediction. An extension could incorporate model-generated feedback or cycle-consistency training while retaining randomly sampled observation masks, although its effect remains to be evaluated.


\section{Experimental Details} \label{app:experimental_details}

\subsection{Reproducibility} \label{app:reproducibility}
We are committed to ensuring the reproducibility of our research. All experimental procedures, data sources, and algorithms used in this study are clearly documented in the paper. The code and datasets will be provided as the supplementary material and be made publicly available upon publication, allowing others to validate our findings and build upon our work.

\subsection{Environments} \label{app:environment}
All models in the experiments of Section~\ref{sec:experiments} are trained for 200 epochs on four NVIDIA B200 GPUs. All inference results for these experiments are obtained using a single NVIDIA B200 GPU. For the additional experiments in Appendix~\ref{app:additional_experiments}, all models are trained for 100 epochs on four NVIDIA A6000 GPUs. Details of the Python version and the specific packages used in the experiments are provided in the environment file included with the accompanying code.

\subsection{Hyperparameters} \label{app:hyperparameter}

In this section, we describe the hyperparameter settings used for each model.
The best hyperparameters of all models were selected based on the validation split, by choosing the configuration that achieves the lowest relative $L_2$ error. For DiffusionPDE and FunDPS, we follow the hyperparameter choices reported in the original papers for model training and architecture and only search the inference-time guidance weight $\zeta$. The candidate guidance weights are those reported in Appendix~\ref{app:hyperparameter_sensitivity}. For SBNO, we use 32 RK4 steps by default, 128 ODE right-hand-side evaluations, corresponding to 256 individual drift-network evaluations. Table~\ref{table:hyperparameters} summarizes the hyperparameters used in all experiments, including the selected guidance weight for each dataset and problem. Unless stated otherwise, all other hyperparameters are fixed across datasets, and the same validation protocol is applied to ensure a fair and reproducible comparison.

\begin{table}[htbp] 
\captionsetup{skip=1.5pt}
\caption{Best hyperparameters for each model used in the main experiments. \label{table:hyperparameters}}
\small
\centering
\resizebox{\linewidth}{!}{%
\begin{tabular}{llcccccccccc}
\toprule
\multirow{2}{*}{Models} &\multirow{2}{*}{Hyperparameter}  & \multicolumn{2}{c}{\textbf{Darcy Flow}} & \multicolumn{2}{c}{\textbf{Poisson}} & \multicolumn{2}{c}{\textbf{Helmholtz}} & \multicolumn{2}{c}{\begin{tabular}[c]{@{}c@{}}\textbf{Nonbounded}\\ \textbf{Navier-Stokes}\end{tabular}} & \multicolumn{2}{c}{\textbf{Gray-Scott}}  \\ 
\cmidrule(lr){3-5} \cmidrule(lr){5-6} \cmidrule(lr){7-8} \cmidrule(lr){9-10}  \cmidrule(lr){11-12} 
& & Forward & Inverse & Forward & Inverse & Forward & Inverse & Forward & Inverse  & Forward & Inverse \\
\midrule
\multirow{4}{*}{Ours} & learning rate & \multicolumn{10}{c}{0.0001}  \\
                      & ODE solver step & \multicolumn{10}{c}{32}  \\
                      & Sigma min     & \multicolumn{2}{c}{0.2} & \multicolumn{2}{c}{0.001} & \multicolumn{2}{c}{0.001} & \multicolumn{2}{c}{0.001}  & \multicolumn{2}{c}{0.001}  \\
                      & Sigma max     & \multicolumn{2}{c}{0.2} &  \multicolumn{2}{c}{0.3} & \multicolumn{2}{c}{0.3} & \multicolumn{2}{c}{0.45} & \multicolumn{2}{c}{0.3}  \\

\midrule
\multirow{2}{*}{FunDPS multi-res}       & Observation guidance  & 1,000 & 25,000  & 5,000 & 10,000 & 10,000 & 5,000 & 5,000 & 1,000  & 5,000 & 10,000  \\
                              & PDE guidance          & 0 & 0 & 0 & 0 & 1 & 1  & 1 & 10  & 0 & 0   \\
\midrule
\multirow{2}{*}{FunDPS}       & Observation guidance  & 5,000 & 50,000  & 5,000 & 20,000 & 15,000 & 1,000 & 5,000 & 1,000 & 10,000 & 10,000  \\
                              & PDE guidance          & 0 & 0 & 0 & 0 & 1 & 1  & 1 & 10  & 0 & 0   \\
\midrule
\multirow{2}{*}{DiffusionPDE} & Observation guidance  & 500  & 1500 & 200 & 300 & 300 & 200 & 500 & 500   & 100 & 100 \\
                              & PDE guidance          & 1000 & 100  & 100 & 1   & 100 & 1 & 100 & 100  & 100 & 100  \\
\midrule
\multirow{3}{*}{FNOPE}    & Layers number     & \multicolumn{10}{c}{6}   \\
                          & Hidden dimension  & \multicolumn{10}{c}{128} \\ 
                          & Head number       & \multicolumn{10}{c}{4} \\ 
\midrule
\multirow{2}{*}{GNOT}     & Layers number     & \multicolumn{10}{c}{4}   \\
                          & Hidden dimension  & \multicolumn{10}{c}{128} \\ 
\midrule
\multirow{2}{*}{FNO}      & Layers number     & \multicolumn{10}{c}{6}   \\
                          & Hidden dimension  & \multicolumn{10}{c}{128} \\
\bottomrule
\end{tabular}%
}
\end{table}

\section{Impact Statement} \label{app:impact_statement}
This work follows standard ethical guidelines for scientific research. SBNO is designed for PDE operator learning under sparse or missing observations and does not involve human-subject data or personal information. We do not see a clear direct path from deploying our method to harmful uses such as surveillance, disinformation, or privacy attacks. We expect that improved forward/inverse PDE modeling can benefit applications like climate and geophysical simulation, biomedical modeling, and engineering analysis where observations are limited. As with most advances in simulation and inverse modeling, there is a general dual-use possibility, so high-stakes deployments should include appropriate validation and governance.

All experiments were conducted in compliance with relevant laws and ethical guidelines, ensuring the integrity of our findings. We are committed to transparency and reproducibility in our research processes. The code will be provided as the supplementary material and be made publicly available upon publication, allowing others to validate our findings and build upon our work.

\clearpage

\clearpage

\end{document}